\documentclass[10pt,twocolumn,letterpaper]{article}

\usepackage{iccv}
\usepackage{times}
\usepackage{epsfig}
\usepackage{graphicx}
\usepackage{amsmath}
\usepackage{amssymb}
\usepackage{widetext}

\usepackage{amsmath,amsfonts,bm}

\def\eqref#1{equation~\ref{#1}}

\def\1{\bm{1}}

\DeclareMathAlphabet{\mathsfit}{\encodingdefault}{\sfdefault}{m}{sl}
\SetMathAlphabet{\mathsfit}{bold}{\encodingdefault}{\sfdefault}{bx}{n}

\usepackage{color,xcolor}
\usepackage{array}
\usepackage{booktabs}
\usepackage{colortbl}
\usepackage{float,wrapfig}
\usepackage{hhline}
\usepackage{multirow}
\usepackage{subcaption} %
\usepackage[font={small}]{caption}

\usepackage{amsmath,amsfonts,amsthm,amssymb}
\usepackage{bm}
\usepackage{nicefrac}
\usepackage{microtype}

\usepackage{changepage}
\usepackage{extramarks}
\usepackage{fancyhdr}
\usepackage{lastpage}
\usepackage{setspace}
\usepackage{soul}
\usepackage{xspace}

\usepackage{balance} %
\usepackage{pdfpages}

\newcolumntype{L}[1]{>{\raggedright\let\newline\\\arraybackslash\hspace{0pt}}m{#1}}
\newcolumntype{C}[1]{>{\centering\let\newline\\\arraybackslash\hspace{0pt}}m{#1}}
\newcolumntype{R}[1]{>{\raggedleft\let\newline\\\arraybackslash\hspace{0pt}}m{#1}}

\newcommand{\sect}[1]{Section~\ref{#1}}

\newcommand{\fig}[1]{Figure~\ref{#1}}
\newcommand{\tbl}[1]{Table~\ref{#1}}

\newcommand{\ignore}[1]{}

\makeatletter
\DeclareRobustCommand\onedot{\futurelet\@let@token\@onedot}
\def\@onedot{\ifx\@let@token.\else.\null\fi\xspace}

\def\eg{e.g\onedot} 
\def\ie{i.e\onedot}

\def\etal{et al\onedot}
\makeatother

\definecolor{MyDarkBlue}{rgb}{0,0.08,1}
\definecolor{MyDarkGreen}{rgb}{0.02,0.6,0.02}
\definecolor{MyDarkRed}{rgb}{0.8,0.02,0.02}
\definecolor{MyDarkOrange}{rgb}{0.40,0.2,0.02}
\definecolor{MyPurple}{RGB}{111,0,255}
\definecolor{MyRed}{rgb}{1.0,0.0,0.0}
\definecolor{MyGold}{rgb}{0.75,0.6,0.12}
\definecolor{MyDarkgray}{rgb}{0.66, 0.66, 0.66}

\newcommand{\modelfull}{Program-Guided Image Manipulator\xspace}
\newcommand{\model}{PG-IM\xspace}
\newcommand{\algorithmbased}{non-learning-based\xspace}

\newcommand{\myparagraph}[1]{\vspace{-14pt}\paragraph{#1}}

\newcommand{\mycell}[1]{\begin{tabular}{@{}l@{}l}#1\end{tabular}}

\usepackage[pagebackref=true,breaklinks=true,letterpaper=true,colorlinks,bookmarks=false]{hyperref}

\iccvfinalcopy %

\def\iccvPaperID{1538} %

\ificcvfinal\fi
\begin{document}

\title{\vspace*{-5pt}Program-Guided Image Manipulators}
\author{
\vspace*{-20pt}
Jiayuan Mao$^{1*}$\qquad Xiuming Zhang$^{1*}$\qquad Yikai Li$^{1,2}$\\
William T. Freeman$^{1,3}$\qquad Joshua B. Tenenbaum$^{1}$\qquad Jiajun Wu$^{1}$\\
\vspace*{3pt}
$^1$MIT CSAIL\qquad $^2$Shanghai Jiao Tong University\qquad $^3$Google Research\\
\vspace*{3pt}
\url{http://pgim.csail.mit.edu}
\vspace*{-10pt}
}

\thispagestyle{empty}

\twocolumn[{%
\renewcommand\twocolumn[1][]{#1}%
\maketitle

\vspace{-15pt}
    \centering
    \includegraphics[width=\textwidth]{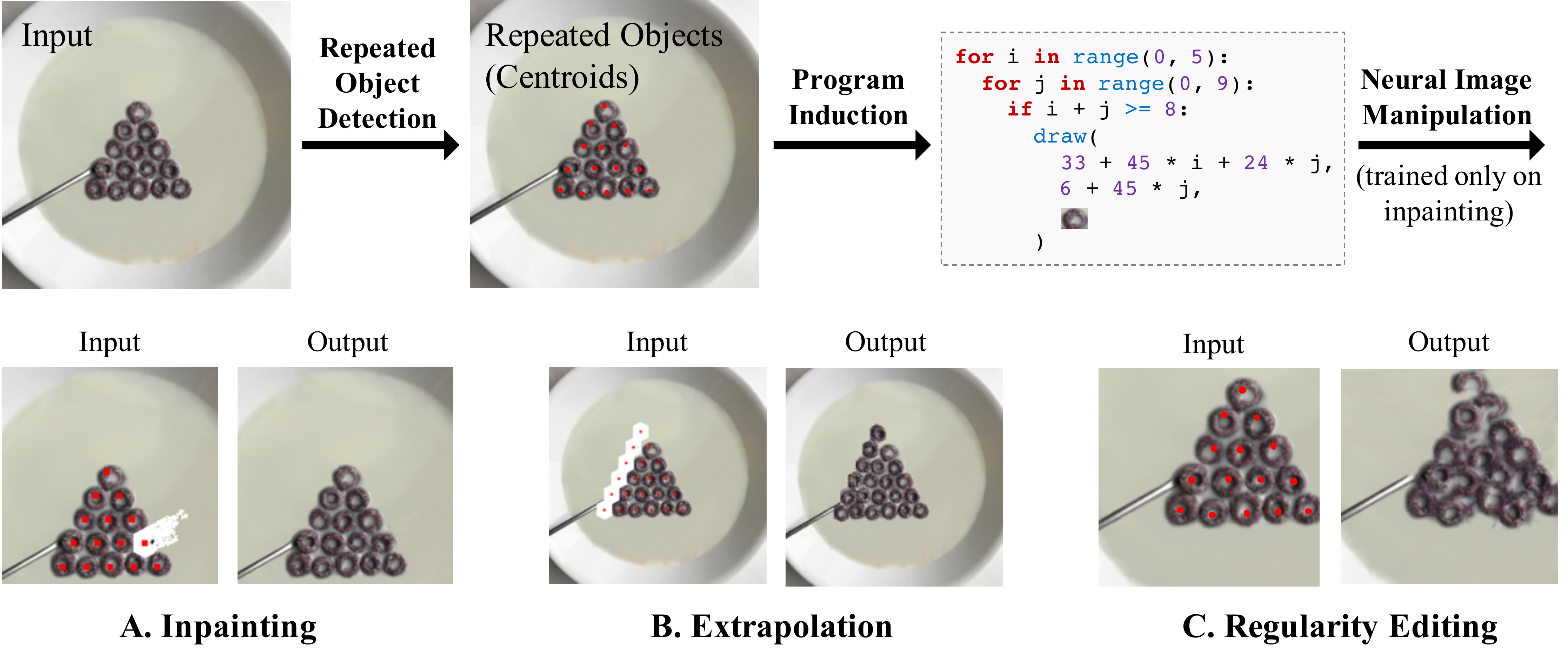}
    \vspace{-25pt}
    \captionof{figure}{Given an input image, the \modelfull (\model) detects repeated entities in the image (pieces of cereal) and then infers a program-like representation for describing the regularity of the image. The regularity representation empowers multiple downstream tasks, such as image inpainting, extrapolation, and regularity editing.}
    \vspace{-10pt}
    \label{fig:teaser}
\vspace{1.2em}
}]
\begin{abstract}
\vspace{-10pt}
Humans are capable of building holistic representations for images at various levels, from local objects, to pairwise relations, to global structures. The interpretation of structures involves reasoning over repetition and symmetry of the objects in the image. In this paper, we present the \modelfull (\model), inducing neuro-symbolic program-like representations to represent and manipulate images. Given an image, \model detects repeated patterns, induces symbolic programs, and manipulates the image using a neural network that is guided by the program. \model learns from a single image, exploiting its internal statistics. Despite trained only on image inpainting, \model is directly capable of  extrapolation and regularity editing in a unified framework. Extensive experiments show that \model achieves superior performance on all the tasks.
\vspace{-20pt}
\end{abstract}
 \footnotetext{$^*$ equal contribution; order determined by a coin toss.}
\thispagestyle{empty}

\vspace{-10pt}
\section{Introduction}
\vspace{-5pt}
Looking at the images in \fig{fig:teaser}, we effortlessly identify the objects (pieces of cereal) in the image, interpret their pairwise relations, and reason over the global {\it regularity}: all pieces of cereal are organized on a 2D lattice with a triangular boundary. This holistic representation empowers our imagination of unseen objects: we can inpaint missing pixels in images, extrapolate images while preserving the regularity~\cite{Rock1990legacy}, and reduce or exaggerate the regularity.

While tremendous progress has been made in object recognition~\cite{He2016Deep} and visual relation detection~\cite{Lu2016Visual}, a global representation for {\it structural regularity} is still missing in these studies. In this paper, we propose to augment deep networks, which are very powerful in pixel-level recognition, with symbolic programs, which are flexible to capture high-level regularity within the image. The intuition is that the disentanglement between perception and reasoning will enable complex image manipulation, preserving both high-level scene structure and low-level object appearance.

Our model, the \modelfull (\model), induces symbolic programs for global regularities and manipulates images with deep generative models guided by the programs. \model consists of three modules: a neural module that detects repeated patterns within the input image, a symbolic program synthesizer that infers programs for spatial regularity (lattice structure) and content regularity (object attributes), and a neural generative model that manipulates images based on the inferred programs.

We demonstrate the effectiveness of \model on two datasets: the Nearly-Regular Pattern dataset \cite{Lettry2017Repeated} and the Facade dataset \cite{Teboul2010Segmentation}. Both datasets contain nearly-regular images with lattice patterns of homogeneous objects. We also extend our experiments to a collection of Internet images with non-lattice patterns and variations in object appearance. Our neuro-symbolic approach robustly outperforms neural and patch-matching-based baselines on multiple image manipulation tasks, such as inpainting, extrapolation, and regularity editing.

\vspace{-10pt}
\section{Related Work}
\vspace{-5pt}
\paragraph{Image manipulation.}

Image manipulation is a long-standing problem in computer vision, graphics, and computational photography, most often studied in the context of image inpainting. Throughout decades, researchers have developed numerous inpainting algorithms operating at various levels of image representations: pixels, patches, and most recently, holistic image features learned by deep  networks. Pixel-based methods often rely on diffusion~\cite{ashikhmin2001synthesizing,Ballester2001Filling} and work well when the holes are small; later, patch-based methods~\cite{Efros2001Image,Barnes2009PatchMatch} accelerate pixel-based methods and achieve better results. Both methods do not perform well in cases that require high-level information beyond background textures.

Deep networks are good at learning semantics from large datasets, and the learned semantic information has been applied to image manipulation~\cite{xie2012image,pathak2016context,ulyanov2018deep}. Many follow-ups have been proposed to improve the results via multi-scale losses~\cite{iizuka2017globally,yang2017high}, contextual attention~\cite{yu2018generative}, partial convolution~\cite{Liu2018Image}, gated convolution~\cite{Yu2018Free}, among others~\cite{Zhou2018Nonstationary,yan2018shift}. Although these methods  achieve impressive inpainting results with the learned semantic knowledge, they  have two limitations: first, they rely on networks to learn object structure implicitly, and may fail to capture explicit, global object structures, such as the round shape of a clock~\cite{xiong2019foreground}; second, the learned semantics is specific to the training set, while real-world test images are likely to be out-of-distribution. Very recently, Xiong~\etal~\cite{xiong2019foreground} and Nazeri~\etal~\cite{nazeri2019edgeconnect} tackled the first problem by explicitly modeling contours to help the inpainting system preserve global object structures. In this paper, we propose to tackle both problems using a combination of bottom-up deep recognition networks and the top-down neuro-symbolic program induction. We apply our approach to scenes with an arbitrary number of objects.

\myparagraph{Program induction and procedural modeling.}

The idea of using procedural modeling for visual data has been a well-studied topic in computer graphics, mostly for indoor scenes~\cite{SymmetryHierarchy,GRAINS,Im2Struct} and 3D shapes~\cite{Li2017GRASS}. More recently, with deep recognition networks, researchers have studied converting 2D images to line-drawing programs~\cite{Ellis2017learning}, primitive sets~\cite{Sharma2018CSGNet}, markup code~\cite{DBLP:journals/corr/DengKR16,beltramelli2017pix2code}, or symbolic programs with attributes~\cite{scene2prog}. These papers tackle synthetic images in a constrained domain, while here we study natural images.

SPIRAL~\cite{SPIRAL} used reinforcement learning to derive ``drawing commands'' for natural images. Their commands are, however, not interpretable, and it is unclear how they can be extended to handle complex relations among a set of objects. Most recently, Young~\etal~\cite{young2019learning} integrated formal representations with deep generative networks and applied it to natural image inpainting. Still, our model differs from theirs in two aspects. First, we use neural modules for discovering repeated patterns in images, which does not require the patch of interest to repeat itself over the entire image (an assumption made in \cite{young2019learning}). Second, their algorithm requires learning semantics on a pre-defined dataset for manipulation (\eg, image extrapolation); in contrast, our model exploits the idea of internal learning~\cite{shocher2018zero} and requires no training data during image manipulation other than the image itself.

\myparagraph{Single-image learning.}

Because visual entropy inside a single image is lower than in a diverse collection of images~\cite{zontak2011internal}, many works have exploited image-level (instead of dataset-level) statistics for various image editing tasks including deblurring~\cite{bahat2017non,michaeli2014blind}, super-resolution~\cite{irani2009super,freedman2011image,Huang2015Single}, and dehazing~\cite{bahat2016blind}. The same philosophy has also been proven successful in deep learning, where neural networks are trained on (and hence overfit to) a single image. Such image-specific networks effectively encode image priors unique to the input image~\cite{ulyanov2018deep}. They can be used for super-resolution~\cite{shocher2018zero}, layer decomposition~\cite{gandelsman2018double}, texture modeling~\cite{bergmann2017learning,Zhou2018Nonstationary}, and even generation tasks~\cite{shaham2019singan,shocher2019ingan}.

Powerful as these approaches are, they often lack a high-level understanding of the input image's global structure (such as the triangular shape formed by the cereal in \fig{fig:teaser}). Consequently, there is usually no guarantee that the original structure gets preserved after the manipulation (e.g., Row 2 of \fig{fig:inpainting}). This work augments single-image learning methods with symbolic reasoning about the input image's global structure, not only providing a natural way of preserving such structure, but also enabling higher-level, semantic manipulation based on the structure (e.g., extrapolating an additional row of cereal following the triangular structure in the teaser figure). %
\begin{figure*}[t]
    \centering
    \vspace{-10pt}
    \includegraphics[width=\textwidth]{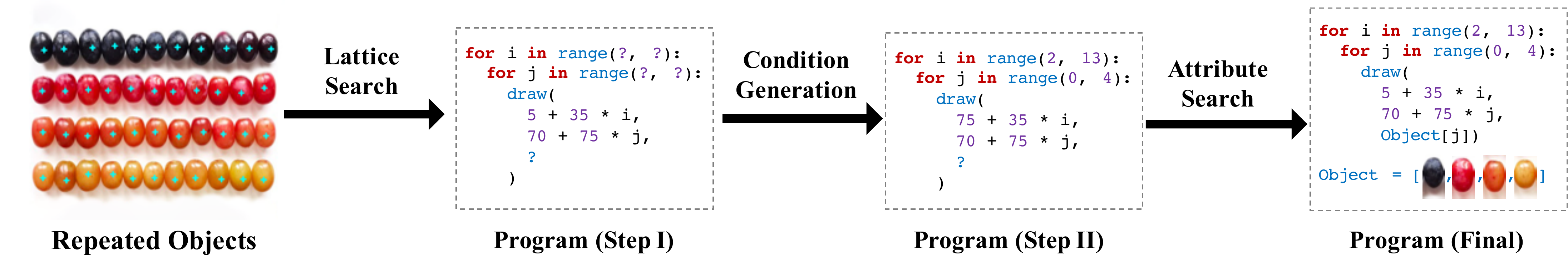}
    \vspace{-25pt}
    \caption{The three-step inference of a program describing the shown repeated pattern. Assuming the input keypoints follow a lattice pattern, we first search for parameters defining the lattice, such as the distance between nearby keypoints and the origin. Next, we fit boundary conditions for the program. Finally, we cluster objects into groups by their visual appearance, and fit an expression describing the variation.}
    \label{fig:pipeline-ps}
    \vspace{-8pt}
\end{figure*}
\vspace{-5pt}
\section{\modelfull}
\vspace{-5pt}

The \modelfull (\model) contains three modules, as shown in \fig{fig:teaser}. First, \model detects repeated objects and make them a variable-length stack (\sect{sec:repeat}). Then, it infers a program to describe the global regularity among the objects (\sect{sec:ps}), with program tokens such as for-loops for repetition and symmetry. Finally, the inferred program facilitates image manipulation, which is performed by a neural painting network (\sect{sec:npn}).

\vspace{-3pt}
\subsection{Repeated Object Detection}
\label{sec:repeat}
\vspace{-3pt}

\model detects repeated objects in the input image with a neural module based on Lettry \etal~\cite{Lettry2017Repeated}. Given the input image, it extracts convolutional feature maps from a pre-trained convolutional neural network (\ie, AlexNet~\cite{Krizhevsky2012Imagenet}). A morphological filter is then applied to the feature maps for extracting activated neurons, resulting in a stack of {\it peakmaps}. Next, assuming the lattice pattern of repeated objects, a voting algorithm is applied to compute the displacements between nearby objects. Finally, an implicit pattern model (IPM) is employed to fit the centroids of objects. Please see~\cite{Lettry2017Repeated} and the supplementary material for details of the algorithm.

\vspace{-3pt}
\subsection{Program Synthesizer}
\label{sec:ps}
\vspace{-3pt}
\begin{figure}[t]
    \centering
    \includegraphics[width=\columnwidth]{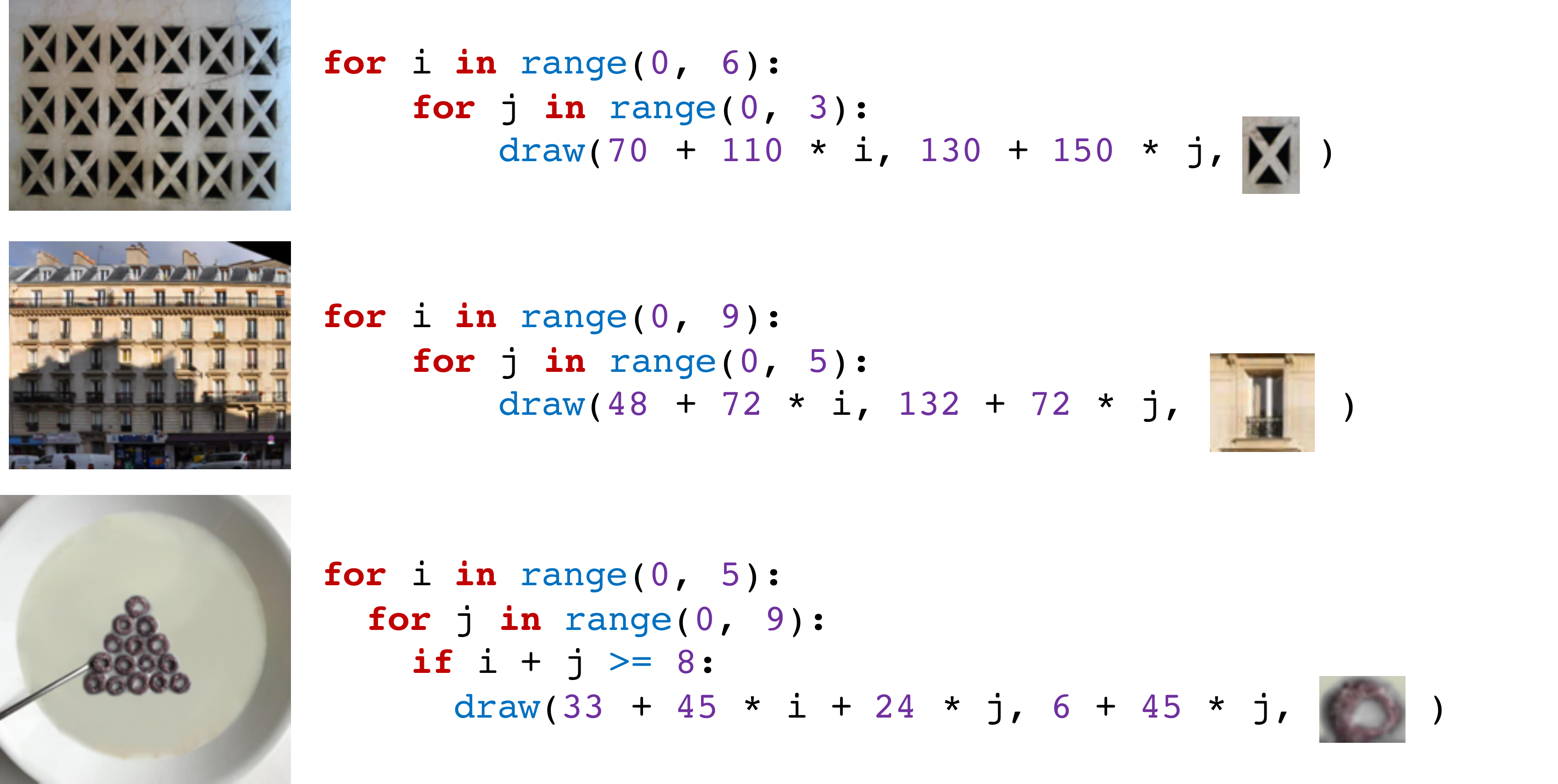}
    \vspace{-15pt}
    \caption{Illustrative programs inferred from (top row) the Nearly-Regular Pattern dataset \cite{Lettry2017Repeated}, (middle row) the Facade dataset \cite{Teboul2010Segmentation}, and (bottom row) Internet images. The DSL of the inferred programs supports for-loops, conditions, and attributes.
    }
    \label{fig:program}
    \vspace{-15pt}
\end{figure}
The program synthesizer takes the centroids of the repeated objects as input and infers a latent program describing the pattern. The input image is partitioned into object patches by constructing a Voronoi graph of all pixels. That is, each pixel is assigned to its nearest centroid, under the metric of Euclidean distance between pixel coordinates. Meanwhile, objects are clustered into multiple groups. When the program reconstructs an object with the {\tt Draw} command, it is allowed to specify both the coordinate of the object's centroid ($x$, $y$) and an integer (namely, the {\it attribute}), indicating which group  this object belongs to. We implement our program synthesizer as a search-based algorithm that finds the {\it simplest} program that reconstructs the pattern.

\begin{table}[t!]
\vspace{-.2em}
\scriptsize
\setlength{\tabcolsep}{2pt}
    \centering
    \begin{tabular}{rcp{0.6\columnwidth}}
    \toprule
        Program & $\longrightarrow$ & For1Stmt \\
        \mycell{For1Stmt} & \mycell{$\longrightarrow$} & \mycell{{\tt For} (~$i$~{\tt in} {\tt range}(Integer, Integer) )\\~~~~\{ For2Stmt \}}\\
        \mycell{For2Stmt} & \mycell{$\longrightarrow$} & \mycell{{\tt For} (~$i$~{\tt in} {\tt range}(Integer, Integer) )\\~~~~\{ CondDrawStmt \}}\\
        CondDrawStmt & $\longrightarrow$ & {\tt If} (Expr $\ge$ 0) \{ CondDrawStmt \}\\
        CondDrawStmt & $\longrightarrow$ & DrawStmt \\
        \mycell{DrawStmt} & \mycell{$\longrightarrow$} & \begin{tabular}{@{}l@{}l}{\tt Draw} (x=Expr, y=Expr,\\~~~~attribute=AttributeExpr )\end{tabular} \\
        AttributeExpr & $\longrightarrow$ & Expr // Integer \\
        AttributeExpr & $\longrightarrow$ & 1 {\tt If} (Expr == 0) else 0 \\
        AttributeExpr & $\longrightarrow$ & 1 {\tt If} (Expr == 0 {\tt and} Expr == 0) else 0 \\
        AttributeExpr & $\longrightarrow$ & 1 {\tt If} (Expr \% Integer == 0) else 0 \\
        AttributeExpr & $\longrightarrow$ & 1 {\tt If} (Expr \% Integer == 0 and Expr \% Integer == 0) else 0 \\
        Expr & $\longrightarrow$ & Integer * $i$ + Integer * $j$ + Integer\\
    \bottomrule
    \end{tabular}
    \vspace{-6pt}
    \caption{The domain-specific language (DSL) for describing image regularities. Language tokens including {\tt For}, {\tt If}, {\tt Integer} and arithmetic/logical operators follow the convention of Python.
    }
    \label{tab:dsl}
        \vspace{-15pt}

\end{table} %
\begin{figure*}[t]
    \centering
    \vspace{-10pt}
    \includegraphics[width=\textwidth]{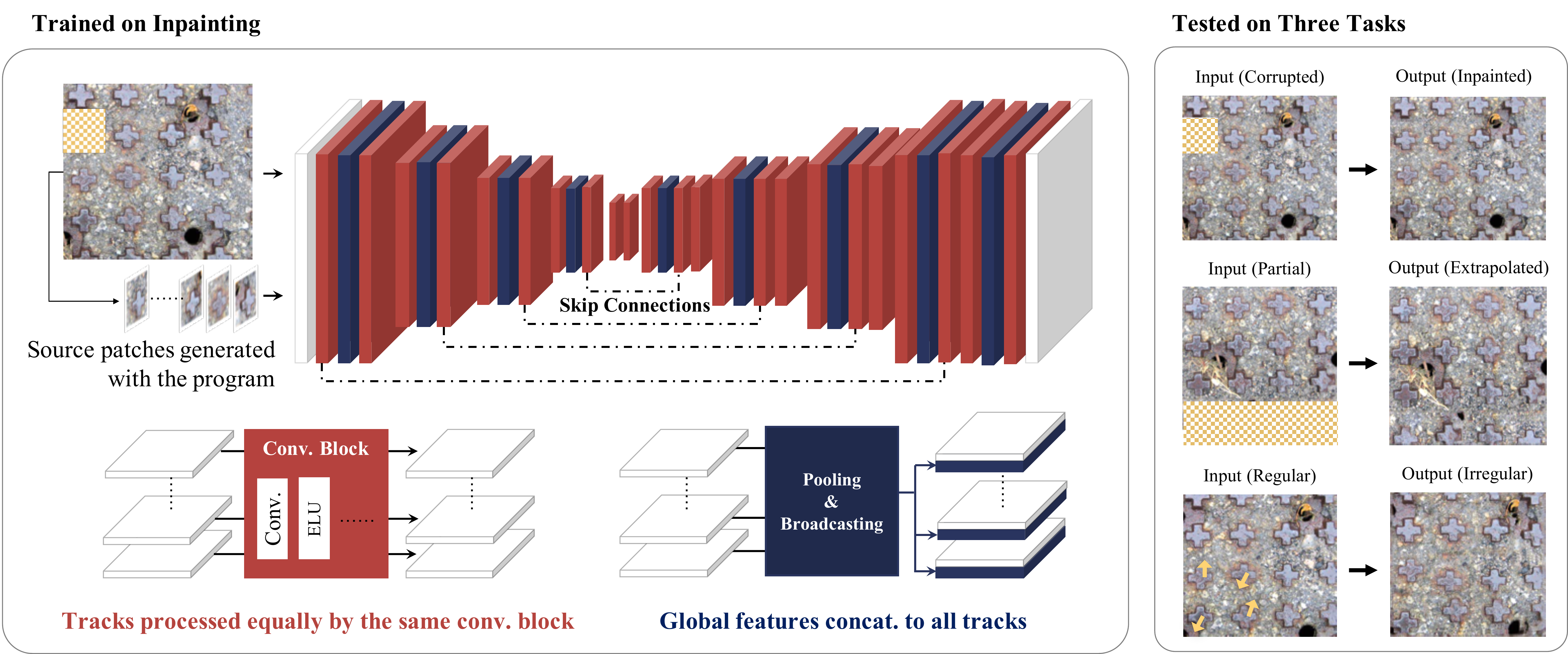}
    \vspace{-15pt}
    \caption{A neural painting network (NPN) takes as input an image and a set of source patches, derived from the image with its program description, and outputs a manipulated image. An NPN learns from a single image, exploiting the image's internal statistics. Trained only on inpainting, it can directly extrapolate and edit the regularity of the input image in a unified inference framework, without any finetuning.}
    \vspace{-15pt}
    \label{fig:pipeline-npn}
\end{figure*}
\myparagraph{Domain-specific language.}

We summarize the domain-specific language (DSL) used by \model for describing object repetition in \tbl{tab:dsl}. In a nutshell, {\tt ForStmt1} and {\tt ForStmt2} jointly define a lattice structure;  {\tt CondDrawExpr} defines the boundary of the lattice; {\tt Draw}  places an object at a given coordinate. {\tt  AttributeExpr} allows the attribute of the object to be conditioned on the loop variables ($i$ and $j$). \fig{fig:program} shows illustrative programs inferred from different datasets.

\myparagraph{Program search.}
Finding the {\it simplest} program for describing a regularity pattern involves  searching over a large compositional space of possible programs, which contains for-loops, if-conditions, coordinate expressions, and attribute expressions. To accelerate the search, we heuristically divides the search process into three steps, as illustrated in \fig{fig:pipeline-ps}. First, we search over all possible expressions for the coordinates, and find the one that fits the detected centroids the best. Second, we determine the conditions (the boundary). Finally, we find the expression for attributes.

\myparagraph{Lattice search.} The lattice search finds the expressions for coordinates $x$ and $y$, ignoring all potential conditions and attribute expressions. Thus, the search process can be simplified as finding a 5-tuple $(b_x, b_y, d_{x,i}, d_{x,j}, d_{y,j})$ that satisfies $x = b_x + i \cdot d_{x,i} + j \cdot d_{x,j}$ and $y = b_y + j \cdot d_{y,j}$.

Each tuple defines a set of centroids $\mathbf{P}$ containing all $(x, y)$ pairs whose coordinates are within the boundary of the whole image. We compare these sets with the centroids $\mathbf{C}$ detected by the repeated pattern detector. We find the optimal tuple as the one that minimizes a cost function
\vspace{-5pt}
\begin{equation}
    \mathcal L_{\text{lat}} = \sum_{(x, y) \in \mathbf{C}} \min\limits_{(u, v) \in \mathbf{P}} \left[ (x-u)^2  + (y - v)^2 \right] + \lambda |\mathbf{P}|,
\end{equation}
where $\lambda = 5$ is a hyperparameter for regularization.
It matches each detected centroid with the nearest one reconstructed by the program. The goal is to minimize the distance between them and a regularization term over the size of $\mathbf{P}$. From a Bayesian inference perspective, $\mathbf{P}$ defines a mixture of Gaussian distribution over the 2D plane. $\mathcal L_{\text{lat}}$ approximates the log-likelihood of the observation $\mathbf{C}$ and a prior distribution over possible $\mathbf{P}$'s, which favors small ones.

\myparagraph{Condition search.} In the next step, we generate the conditions of the program, assuming all centroids fit in a convex hull. This assumption covers both rectangular lattices and triangular lattices (see \fig{fig:program} for examples). Since all pairs $(x, y)$ are computed by an affine transformation of all $(i, j)$'s, the conditions can be determined by computing the convex hull of all $(i, j)$'s that are matched with detected centroids.

Specifically, we first match each coordinate in $\mathbf{P}$ with $\mathbf{C}$ by computing a minimum cost assignment between two sets, where the distance metric is the Euclidean distance in the 2D coordinate space. We then find the convex hull of all assigned pairs $(i, j)$. We use the boundary of the convex hull as the conditions. The conditions include the boundary conditions of for-loops as well as optional if-conditions.

\myparagraph{Attribute search.} The last step is to find the expression that best describes the variance in object appearance (\ie, their attributes).
Attributes are represented as a set of integers. Instead of clustering, we assign discrete labels to individual patches. The label of the patch in row $p_i$, column $p_j$ is a function of $(p_i, p_j)$. Shown in \tbl{tab:dsl}, each possible {\tt AttributeExpr} defines an attribute assignment function $A(p) \triangleq A(p_i, p_j)$ for all centroids $p = (p_i, p_j) \in \mathbf{P}$.
We say an expression fits the image if patches of the same label share similar visual appearance. Formally, we find the optimal parameters for the attribute expression that minimizes
\vspace{-10pt}
\begin{equation}
    \mathcal L_{\text{attr}}=\sum_{p \in \mathbf{P}} \sum_{q \in \mathbf{P}} \big( \text{sgn}(A(p), A(q)) \cdot \textit{d}(p, q) \big) + \mu |A(\mathbf{P})|,
\end{equation}
where $\text{sgn}(A(p), A(q))=1$ if $A(p)=A(q)$, and $-1$ otherwise. $\textit{d}$ computes the pixel-level difference between two patches centered at $(p_i, p_j)$ and $(q_x, q_y)$, respectively. $\mu = 10$ is a scalar hyperparameter of the regularization strength. $|A(\mathbf{P})|$ computes the number of distinct values of $A(p)$ for all $p \in \mathbf{P}$.
The inference is done by searching over possible integer templates (\eg, $a i + b j + c$) and binary templates (\eg, $(a i + b j + c~{\text{//}}~d~\%~e) == 0$), and the coefficients ($a, b, c, \dots$).

\vspace{-5pt}
\subsection{Neural Painting Networks}
\vspace{-5pt}
\label{sec:npn}

We propose the neural painting network (NPN), a neural architecture for manipulating images with the guidance of programs. It unifies three tasks: inpainting, extrapolation, and regularity editing in a single framework. The key observation is that all three tasks can be cast as filling pixels in images. For illustrative simplicity, we first consider the task of inpainting missing pixels in the image, and then discuss how to perform extrapolation and regularity editing using the same inpainting-trained network.

\myparagraph{Patch aggregation.} We first aggregate all pixels from other ``objects'' (loosely defined by the induceted program) to inpaint the missing pixels. Denote all object centroids reconstructed by the program as $\mathbf{P}$, the centroid of the object patch containing missing pixels $(x_0, y_0)$, and all other centroids $\mathbf{P}^- = \mathbf{P} \setminus \{(x_0, y_0)\}$. The aggregation is performed by generating $|\mathbf{P}^-|$ images, the $i$-th of which is obtained by translating the original image such that the centroid of the $i$-th object in $\mathbf{P}^-$ is centered at $(x_0, y_0)$. Pixels without a value after the shift are treated as 0. We stack the input image with missing pixels plus all the $|\mathbf{P}^-|$ images (the ``patch source'') as the input to the network.

\myparagraph{Architecture.} Our neural painting network (NPN) has a U-Net~\cite{Ronneberger2015U} encoder-decoder architecture, designed to handle a variable number of input images and be invariant to their ordering. Demonstrated in \fig{fig:pipeline-npn}, the network contains a stack of shared-weight convolution blocks and max-pooling layers that aggregate information across all inputs.
Paired downsampling and upsampling layers (convolution layers with strides) are skip-connected. The input of the network is the stack of the corrupted input image plus source patches, and the output of the network is the inpainted image. A detailed printout of the generator's architecture can be found in the supplemental document.

The key insight of our design of the NPN is that it handles a variable number of input images in any arbitrary order. To this end,
inspired by Aittala~\etal~\cite{Aittala2018Burst} and Qi~\etal~\cite{Qi2017PointNet}, we have a single encoder-decoder that processes the $|\mathbf{P}^-|+1$ images equally (``tracks''), and the intermediate feature maps from these tracks get constantly max-pooled into a ``global'' feature map, which is then broadcast back to the $|\mathbf{P}^-|+1$ tracks and concatenated to each track's local feature map to be processed by the next block. Intuitively, the network is guided to produce salient feature maps that will ``survive'' the max-pooling, and the tracks exchange information by constantly absorbing the global feature map.

\myparagraph{Extrapolation and regularity editing as recurrent inpainting.}

A key feature of program-guided NPNs is that although they are trained only on the inpainting task, they are able to be used directly for image extrapolation and regularity editing. With the program description of the image, NPNs are aware of where the entities are in the image, and hence able to cast extrapolation as recurrent inpainting of multiple corrupted objects. For instance, to extrapolate a 64-pixel wide stripe to the right, an NPN first queries the program description for where the new peaks are, and then recurrently inpaints each object given all the previously inpainted ones. Similarly for image regularity editing, when the (regularly spaced) centroids provided by the program get randomly perturbed, the pixels falling into their Voronoi cells move together with them accordingly, leaving many ``cracks'' on the image, which the NPN then inpaints recurrently.

\myparagraph{Training.} We train our NPNs with the same training paradigm as Isola~\etal~\cite{Isola2017Image}. We compute an L1 loss and a patch-based discriminator loss, between the generated (inpainted) image and the ground-truth image. We train image-specific NPNs for each individual image in the dataset. While only training the network to inpaint missing pixels, we show that the network can perform other tasks such as image extrapolation and regularity editing, by only changing the input to the network during inference. Other implementation details such as the hidden dimensions, convolutional kernel sizes, and training hyperparameters can be found in the supplementary material.
\vspace{-5pt}
\section{Experiments and Applications}
\vspace{-5pt}
\begin{figure*}[thbp]
    \centering
    \includegraphics[width=0.92\textwidth]{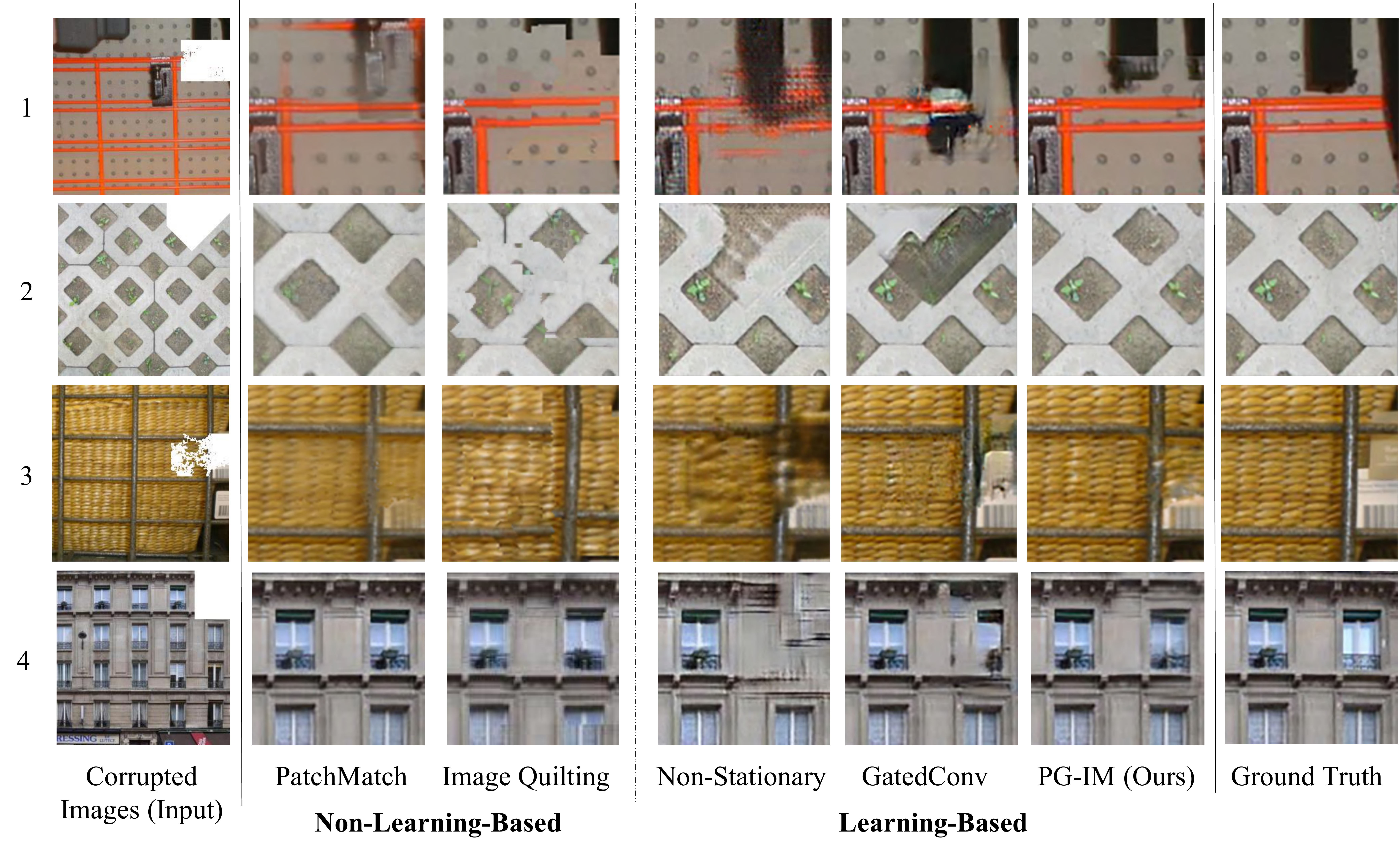}
    \vspace{-15pt}
    \caption{Corrupted input images and inpainting results (zoomed-in) by \model and the baselines. The white pixels in the leftmost column are missing pixels to inpaint. The rightmost column shows the ground-truth patches. \model inpaints realistic image patches that are consistent with the intricate global regularity and meanwhile different from the original, ground-truth patches.}
    \vspace{-15pt}
    \label{fig:inpainting}
\end{figure*}
We provide both quantitative and qualitative comparisons with the baselines on two standard image manipulation tasks: inpainting and extrapolation. We also show the direct application of our approach to image regularity editing, a task where the regularity of an image's global structure gets exaggerated or reduced. It is worth mentioning that these three problems can be solved with a single model trained for inpainting (see~\sect{sec:npn} for details). Finally, we demonstrate how our program induction easily incorporates object attributes (\eg, colors) in Internet images, in turn enabling our NPNs to manipulate images with high-level reasoning in an attribute-aware fashion. Please see the supplemental material for ablation studies that evaluate each major component of \model. We start with an introduction to the datasets and baseline methods we consider.

\vspace{-3pt}
\subsection{Dataset}
\vspace{-3pt}
We compare the performance of \model with other baselines on two datasets: the Nearly-Regular Pattern (NRP) dataset \cite{Lettry2017Repeated} and the Facade dataset \cite{Teboul2010Segmentation}. The Nearly-Regular Pattern dataset contains a collection of 48 rectified images with a grid or nearly grid repetition structure. The Facade dataset, specifically the CVPR 2010 subset, contains 109 rectified images of facades.

\vspace{-3pt}
\subsection{Baselines}
\vspace{-3pt}
We consider two groups of baseline methods: non-learning-based and learning-based. Among the non-learning-based methods are Image Quilting~\cite{Efros2001Image} and PatchMatch~\cite{Barnes2009PatchMatch}, both of which are based on the stationary assumption of the image structure. Intuitively, to inpaint a missing pixel, they fill it with the content of another existing pixel with the most similar context. Being unaware of the objects in the image, they rely on human-specified hyperparameters, such as the context window size, to produce reliable results. More importantly, in the case of extrapolation, the user needs to specify which pixels to paint, implicitly conveying the concept of objects to the algorithms. For PatchMatch and Image Quilting, we search for one set of optimal hyperparameters and apply that to the entire test set.

We also compare \model with a learning-based, off-the-shelf algorithm for image inpainting: GatedConv \cite{Yu2018Free}. They use neural networks for inpainting missing pixels by learning from a large-scale dataset (Place365 \cite{zhou2017places}) of natural images. GatedConv is able to generate novel objects that do not appear in the input image, which is useful for semantic photo editing. However, this may not be desired when the image of interest contains repeated but \emph{unique} patterns: although a pattern appears repeatedly in the image of interest, it may not appear anywhere else in the dataset.
\begin{table}[t!]
\small
    \centering\vspace{5pt}
    \begin{tabular}{lcc}
        \toprule
        Method & L1 Mean (Std.) & Inception Score \\
        \midrule
        \multicolumn{3}{c}{{\bf Nearly-Regular Patterns}~\cite{Lettry2017Repeated}}\\
        \midrule
        Image Quilting~\cite{Efros2001Image} & {\bf 12.30 (2.903)} & {\bf 1.253}\\
        PatchMatch~\cite{Barnes2009PatchMatch} & 83.91 (17.62) & 1.210\\ \midrule
        GatedConv~\cite{Yu2018Free} & 50.45 (16.46) & 1.196\\
        Non-Stationary~\cite{Zhou2018Nonstationary} & 103.7 (23.87) & 1.186\\
        \model~(ours) & {\bf 21.48} ({\bf 5.375}) & {\bf 1.229}\\
        \midrule
        \multicolumn{3}{c}{{\bf Facade}~\cite{Teboul2010Segmentation}}\\
        \midrule
        Image Quilting~\cite{Efros2001Image} & {\bf 13.50 (6.379)} & 1.217\\
        PatchMatch~\cite{Barnes2009PatchMatch} & 81.35 (25.28) & {\bf 1.219}\\ \midrule
        GatedConv~\cite{Yu2018Free} & 26.26 (133.9) & 1.186\\
        Non-Stationary~\cite{Zhou2018Nonstationary} & 133.9 (39.75) & 1.199\\
        \model~(ours) & {\bf 14.40} ({\bf 7.781}) & {\bf 1.218}\\
        \bottomrule
    \end{tabular}
    \vspace{-5pt}
    \caption{We compare \model against off-the-shelf neural baselines for image inpainting on both datasets. Our method outperforms neural baselines with a remarkable margin across all metrics.
    }
    \vspace{-20pt}
    \label{tab:inpainting-number}
\end{table}
Therefore, we also consider another learning-based baseline, originally designed for image extrapolation: Non-Stationary Texture Synthesis (Non-Stationary) \cite{Zhou2018Nonstationary}. In their framework, an image-specific neural network is trained for each input image. Its objective is to extrapolate a small (usually unique) patch ($k \times k$) into a large one ($2k\times 2k$). Although both of their method and \model use single-image training for generating missing pixels, \model uses symbolic programs as the guidance of the networks, enjoying both interpretability and better performance for complex structures. We also implement a variant of Non-Stationary, which keeps the neural architecture and training paradigm as the original version for texture synthesis, but use the same inpainting data as our method for inpainting. For a fair comparison, we train Non-Stationary and \model with single sets of optimal hyperparameters on all test images. For more results and analysis, please refer to the supplementary material.

\begin{figure*}[thbp]
    \centering
    \includegraphics[width=0.83\textwidth]{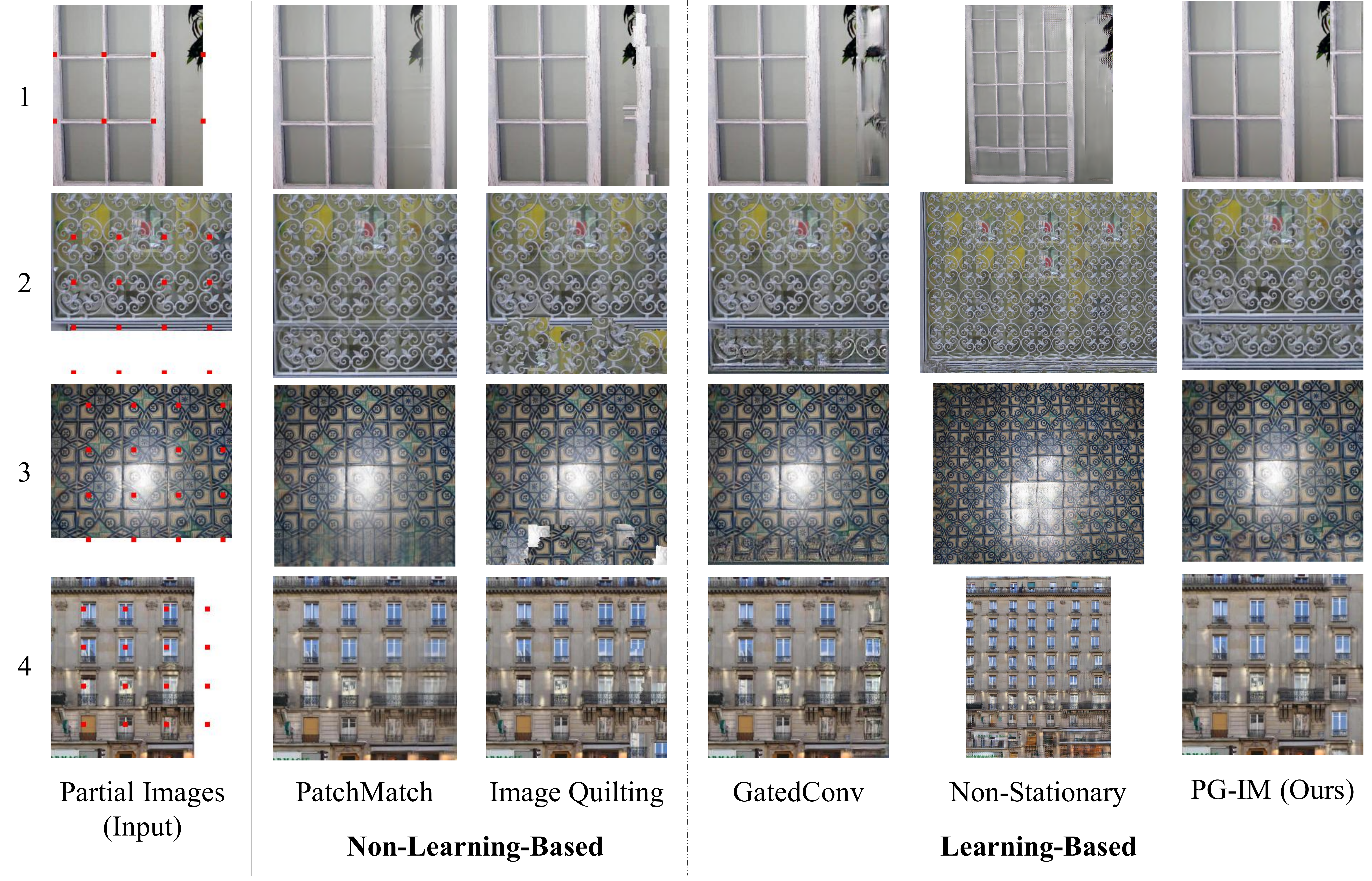}
    \vspace{-10pt}
    \caption{Extrapolation results by \model and the baselines. The white pixels in the leftmost column indicate the pixels to be extrapolated. \model generates realistic images while preserving global regularity. In contrast, GatedConv fails to capture the regularity; Non-Stationary does not preserve the original image contents; PatchMatch tends to generate blurry images in smoothing the transition; Image Quilting does not guarantee the global structure gets preserved.}
    \vspace{-15pt}
    \label{fig:extrapolation}
\end{figure*}
\vspace{-3pt}
\subsection{Inpainting}
\vspace{-3pt}

We compare \model with GatedConv, Image Quilting, and PatchMatch on the task of image inpainting. For quantitative evaluations, we use the NRP and Facade datasets, each of whose images gets randomly corrupted 100 times, giving us a total of around 15,000 test images.

\tbl{tab:inpainting-number} summarizes the quantitative scores of different methods. Following \cite{Liu2018Image}, we compare the L1 distance between the inpainted image and the original image, as well as Inception score (IS) of the inpainted image. For all the approaches, we hold out a test patch whose pixels are never seen by the networks during training, and use that patch for testing. Quantitatively, \model outperforms the other learning-based methods by large margins across both datasets in both metrics. \model recovers missing pixels a magnitude more faithful to the ground-truth images than Non-Stationary in the L1 sense. It also has a small variance across different images and input masks. For comparisons with non-learning-based methods, although Image Quilting achieves the best L1 score, it tends to break structures in the images, such as lines and grids (see  \fig{fig:inpainting} for such examples). Note that the reason why PatchMatch has worse L1 scores is that it also modifies pixels around the holes to achieve better image-level consistency. In contrast, the other methods including PG-IM only inpaint holes and modify nothing else in the images.

\begin{figure*}[thbp]
    \centering
    \includegraphics[width=\textwidth]{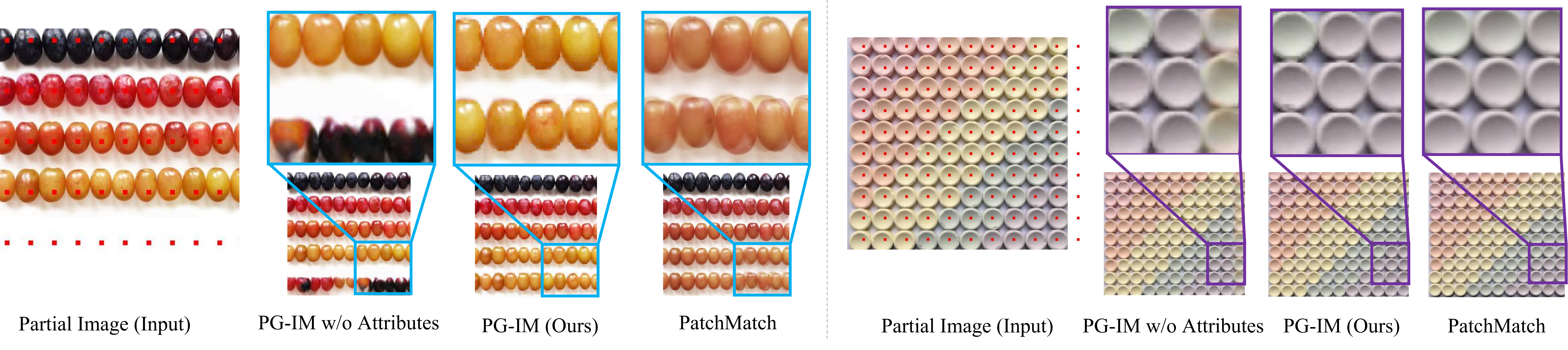}
    \vspace{-20pt}
    \caption{\model can reason about the attribute regularity of images, which supports object appearance--aware image extrapolation. \model w/o Attributes denotes a variant of \model that does not include attributes. See the main text for detailed analysis and comparison.}
    \vspace{-15pt}
    \label{fig:attribute}
\end{figure*}

Qualitative results for inpainting are presented in~\fig{fig:inpainting}.
Overall, our approach is able to preserve the ``objects'' in the test images even if the objects are completely missing, while other learning-based approaches either miss the intricate structures (Non-Stationary on Images 1 and 2), or produce irrelevant patches (learned from largely diverse image datasets) that break the global structure of this particular image (\eg, GatedConv on Image 2). Note how the image patches inpainted by our approach is realistic and meanwhile quite different from the ground-truth patches (compare our inpainting with the ground-truth Image 4). For the non-learning-based approaches, the baselines suffer from blurry outputs and sometimes produce inconsistent connections to the original image on boundaries. Moreover, as we demonstrate in~\fig{fig:attribute}, unlike our approach that combines high-level symbolic reasoning and lower-level pixel manipulations, PatchMatch fails to manipulate the pixels in an attribute-aware fashion.

Runtime-wise, learning-methods including PG-IM, once trained, inpaint an image in a forward pass (around 100ms on GPUs), whereas \algorithmbased approaches take around 15 minutes to inpaint one image.

\vspace{-3pt}
\subsection{Extrapolation}
\vspace{-3pt}

\fig{fig:extrapolation} shows the extrapolation results by \model and the baselines. With the program description of the images, \model naturally knows \emph{where} to extrapolate to, \eg, by incrementing the for-loop range. This contrasts with the baselines that either require the user to specify which pixels to extrapolate (PatchMatch, Image Quilting, and GatedConv), or simply extrapolate to every possible direction (Non-Stationary). Knowing where to extrapolate is particularly crucial for images where the objects do not scatter all over. Take the pieces of cereal in~\fig{fig:teaser}B as an example. \model reasons about the global structure that the pieces of cereal form, decides where to extrapolate to by relaxing its program conditions, and finally extrapolates a new row.

As PatchMatch greedily ``copies from'' patches with the most similar context, certain pixels may come from different patches, therefore producing blurry extrapolation results (Images 1, 3, and 4). Learning from large-scale image datasets, GatedConv fails to capture the repeated patterns specific to each individual image, thus generating patterns that do not connect to the image boundary consistently (Images 2 and 3). Non-Stationary treats the entire image as consisting of only patterns of interest and expands the texture along all four directions; artifacts show up when the image contains more than the texture (bottom of Image 4). Also interesting is that Non-Stationary can be viewed as a super-resolution algorithm, in the sense that it is interpolating among the replicated objects. As the rightmost column shows, during extrapolation, \model produces realistic and sharp patches (Image 1), preserves the images' global regularity, and connects consistently to the image boundary (Images 2-4).

\begin{figure}[t]
    \centering
    \includegraphics[width=\columnwidth]{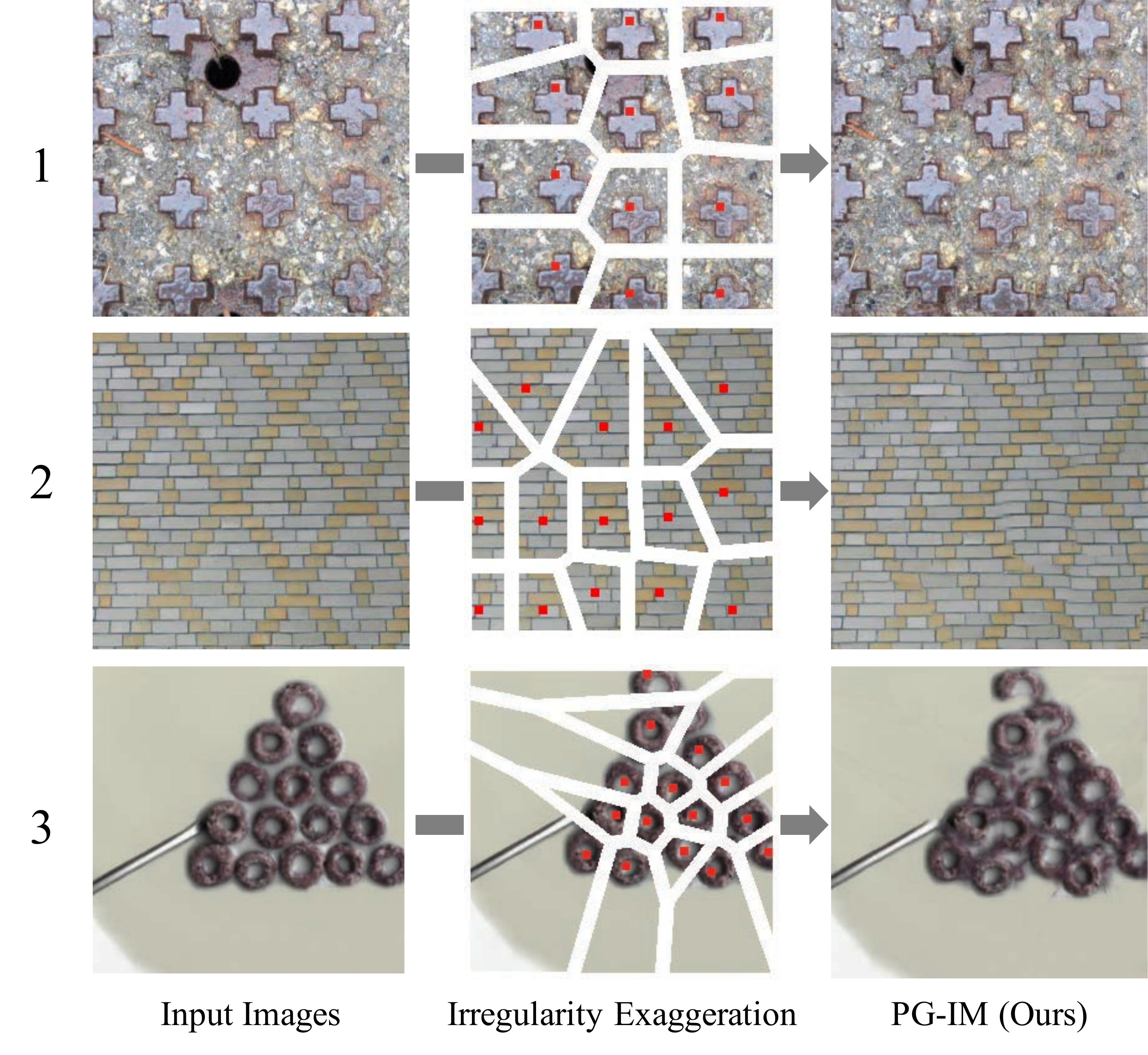}
    \vspace{-20pt}
    \caption{\model enables automated and semantic-aware irregularity exaggeration. By comparing the centroids of the detected objects and the ones reconstructed by the program, we can measure and exaggerate the structural irregularity of input images.}
    \vspace{-15pt}
    \label{fig:editing}
\end{figure}
\vspace{-3pt}
\subsection{Image Regularity Editing}
\vspace{-3pt}

With a program describing the image's \emph{ideal} global regularity, \model is able to exaggerate imperfections in the global regularity by magnifying the discrepancy between what the program depicts and the detected object centroids. A similar task has been discussed by~\cite{Dekel2015Revealing}. In~\fig{fig:editing}, we magnify the displacement vectors between the program-provided and detected centroids by two, and shift the Voronoi cells together with their respective centroids, leaving missing values among the cells. An NPN then fills in the gaps by recurrent inpainting.

\vspace{-3pt}
\subsection{Attribute Regularity}
\vspace{-3pt}

Beyond using for-loops and if-conditions to capture the global regularity of objects, \model can also reason about the regularity of object appearance variations (\ie, the {\it attribute} regularity). Our model automatically clusters objects into groups. Beyond knowing where to extrapolate to, with the attribute regularity described by the program, our NPNs generate new pixels from only patches of the correct attributes.

\fig{fig:attribute} illustrates this idea. We show the image extrapolation results on images with attribute regularities, and compare \model with a variant that does not consider object attributes, as well as a strong baseline: PatchMatch. Without explicit modeling of object attributes, the color of the new objects generated by \model w/o Attributes fails to preserve the global attribute regularity. Meanwhile, due to the existence of objects with similar colors, PatchMatch mixes up two different colors, resulting in blurry output patches (\fig{fig:attribute}L) or  extrapolation results that break the global attribute regularity (the middle object in the top row of \fig{fig:attribute}R's zoom-in windows should be purple, not green).

\vspace{-5pt}
\section{Discussion}
\vspace{-5pt}

This paper presents a neuro-symbolic approach to describing and manipulating natural images with repeated patterns. It combines the power of program induction---as symbolic tools for describing  repetition, symmetry, and attributes---and deep neural networks---as powerful image generative models. \model support various tasks: image inpainting, extrapolation, and regularity editing.

Our results also suggest multiple future  directions.
First, the variations in object appearance are currently handled as discrete properties. We leave the interpretation of attributes that have continuous values, such as the color spectrum in \fig{fig:attribute}, as future work.
Second, given only a facade image containing a number of windows, humans can extrapolate the image by adding doors at the bottom and roof at the top. Combining regularity inference and data-driven approaches is a meaningful direction.
Finally, the representational power of \model is limited by the DSL. \model currently does not generalize to unseen patterns, such as rotational patterns. Future works may consider a more flexible DSL, or even discovering new patterns from data.

\noindent {\bf Acknowledgements.} We thank Michal Irani for helpful discussions and suggestions. This work is supported by the Center for Brains, Minds and Machines (NSF \#1231216), NSF \#1447476, ONR MURI N00014-16-1-2007, IBM Research, and Facebook.
{\small
\bibliographystyle{ieee_fullname}
\bibliography{main,reference}

\begin{thebibliography}{10}\itemsep=-1pt

\bibitem{Aittala2018Burst}
Miika Aittala and Fr{\'e}do Durand.
\newblock Burst image deblurring using permutation invariant convolutional
  neural networks.
\newblock In {\em ECCV}, 2018.

\bibitem{ashikhmin2001synthesizing}
Michael Ashikhmin.
\newblock Synthesizing natural textures.
\newblock In {\em I3D}, 2001.

\bibitem{bahat2017non}
Yuval Bahat, Netalee Efrat, and Michal Irani.
\newblock Non-uniform blind deblurring by reblurring.
\newblock In {\em ICCV}, 2017.

\bibitem{bahat2016blind}
Yuval Bahat and Michal Irani.
\newblock Blind dehazing using internal patch recurrence.
\newblock In {\em ICCP}, 2016.

\bibitem{Ballester2001Filling}
Coloma Ballester, Marcelo Bertalmio, Vicent Caselles, Guillermo Sapiro, and
  Joan Verdera.
\newblock Filling-in by joint interpolation of vector fields and gray levels.
\newblock {\em IEEE TIP}, 10(8):1200--1211, 2001.

\bibitem{Barnes2009PatchMatch}
Connelly Barnes, Eli Shechtman, Adam Finkelstein, and Dan Goldman.
\newblock Patchmatch: a randomized correspondence algorithm for structural
  image editing.
\newblock {\em ACM TOG}, 28(3):24, 2009.

\bibitem{beltramelli2017pix2code}
Tony Beltramelli.
\newblock Pix2code: Generating code from a graphical user interface screenshot.
\newblock In {\em ACM SIGCHI Symposium on Engineering Interactive Computing
  Systems}, EICS, 2018.

\bibitem{bergmann2017learning}
Urs Bergmann, Nikolay Jetchev, and Roland Vollgraf.
\newblock Learning texture manifolds with the periodic spatial gan.
\newblock In {\em ICML}, 2017.

\bibitem{Dekel2015Revealing}
Tali Dekel, Tomer Michaeli, Michal Irani, and William~T Freeman.
\newblock Revealing and modifying non-local variations in a single image.
\newblock {\em ACM TOG}, 34(6):227, 2015.

\bibitem{DBLP:journals/corr/DengKR16}
Yuntian Deng, Anssi Kanervisto, Jeffrey Ling, and Alexander~M Rush.
\newblock Image-to-markup generation with coarse-to-fine attention.
\newblock In {\em ICML}, 2017.

\bibitem{Efros2001Image}
Alexei~A Efros and William~T Freeman.
\newblock Image quilting for texture synthesis and transfer.
\newblock In {\em CGIT}, 2001.

\bibitem{Ellis2017learning}
Kevin Ellis, Daniel Ritchie, Armando Solar-Lezama, and Josh Tenenbaum.
\newblock Learning to infer graphics programs from hand-drawn images.
\newblock In {\em NeurIPS}, 2018.

\bibitem{freedman2011image}
Gilad Freedman and Raanan Fattal.
\newblock Image and video upscaling from local self-examples.
\newblock {\em ACM TOG}, 30(2):12, 2011.

\bibitem{gandelsman2018double}
Yossi Gandelsman, Assaf Shocher, and Michal Irani.
\newblock ``{D}ouble-{DIP}'': Unsupervised image decomposition via coupled
  deep-image-priors.
\newblock In {\em CVPR}, 2019.

\bibitem{SPIRAL}
Yaroslav Ganin, Tejas Kulkarni, Igor Babuschkin, S.~M.~Ali Eslami, and Oriol
  Vinyals.
\newblock Synthesizing programs for images using reinforced adversarial
  learning.
\newblock In {\em ICML}, 2018.

\bibitem{irani2009super}
Daniel Glasner, Shai Bagon, and Michal Irani.
\newblock Super-resolution from a single image.
\newblock In {\em ICCV}, 2009.

\bibitem{He2016Deep}
Kaiming He, Xiangyu Zhang, Shaoqing Ren, and Jian Sun.
\newblock Deep residual learning for image recognition.
\newblock In {\em CVPR}, 2016.

\bibitem{Huang2015Single}
Qixing Huang, Hai Wang, and Vladlen Koltun.
\newblock Single-view reconstruction via joint analysis of image and shape
  collections.
\newblock {\em ACM TOG}, 34(4):87, 2015.

\bibitem{iizuka2017globally}
Satoshi Iizuka, Edgar Simo-Serra, and Hiroshi Ishikawa.
\newblock Globally and locally consistent image completion.
\newblock {\em ACM TOG}, 36(4):107, 2017.

\bibitem{Isola2017Image}
Phillip Isola, Jun-Yan Zhu, Tinghui Zhou, and Alexei~A Efros.
\newblock Image-to-image translation with conditional adversarial networks.
\newblock In {\em CVPR}, 2017.

\bibitem{Krizhevsky2012Imagenet}
Alex Krizhevsky, Ilya Sutskever, and Geoffrey~E Hinton.
\newblock Imagenet classification with deep convolutional neural networks.
\newblock In {\em NeurIPS}, 2012.

\bibitem{Lettry2017Repeated}
Louis Lettry, Michal Perdoch, Kenneth Vanhoey, and Luc Van~Gool.
\newblock Repeated pattern detection using cnn activations.
\newblock In {\em WACV}, 2017.

\bibitem{Li2017GRASS}
Jun Li, Kai Xu, Siddhartha Chaudhuri, Ersin Yumer, Hao Zhang, and Leonidas
  Guibas.
\newblock Grass: Generative recursive autoencoders for shape structures.
\newblock In {\em SIGGRAPH}, 2017.

\bibitem{GRAINS}
Manyi Li, Akshay~Gadi Patil, Kai Xu, Siddhartha Chaudhuri, Owais Khan, Ariel
  Shamir, Changhe Tu, Baoquan Chen, Daniel Cohen-Or, and Hao Zhang.
\newblock Grains: Generative recursive autoencoders for indoor scenes.
\newblock {\em ACM TOG}, 38(2):12:1--12:16, 2019.

\bibitem{Liu2018Image}
Guilin Liu, Fitsum~A. Reda, Kevin~J. Shih, Ting-Chun Wang, Andrew Tao, and
  Bryan Catanzaro.
\newblock Image inpainting for irregular holes using partial convolutions.
\newblock In {\em ECCV}, 2018.

\bibitem{scene2prog}
Yunchao Liu, Zheng Wu, Daniel Ritchie, William~T. Freeman, Joshua~B. Tenenbaum,
  and Jiajun Wu.
\newblock Learning to describe scenes with programs.
\newblock In {\em ICLR}, 2019.

\bibitem{Lu2016Visual}
Cewu Lu, Ranjay Krishna, Michael Bernstein, and Li Fei-Fei.
\newblock Visual relationship detection with language priors.
\newblock In {\em ECCV}, 2016.

\bibitem{michaeli2014blind}
Tomer Michaeli and Michal Irani.
\newblock Blind deblurring using internal patch recurrence.
\newblock In {\em ECCV}, 2014.

\bibitem{nazeri2019edgeconnect}
Kamyar Nazeri, Eric Ng, Tony Joseph, Faisal Qureshi, and Mehran Ebrahimi.
\newblock Edgeconnect: Generative image inpainting with adversarial edge
  learning.
\newblock {\em arXiv:1901.00212}, 2019.

\bibitem{Im2Struct}
Chengjie Niu, Jun Li, and Kai Xu.
\newblock {Im2Struct: Recovering 3D Shape Structure from a Single RGB Image}.
\newblock In {\em CVPR}, 2018.

\bibitem{pathak2016context}
Deepak Pathak, Philipp Krahenbuhl, Jeff Donahue, Trevor Darrell, and Alexei~A
  Efros.
\newblock Context encoders: Feature learning by inpainting.
\newblock In {\em CVPR}, 2016.

\bibitem{Qi2017PointNet}
Charles~R Qi, Li Yi, Hao Su, and Leonidas~J Guibas.
\newblock Pointnet++: Deep hierarchical feature learning on point sets in a
  metric space.
\newblock In {\em NeurIPS}, 2017.

\bibitem{Rock1990legacy}
Irvin Rock and Stephen Palmer.
\newblock The legacy of gestalt psychology.
\newblock {\em Sci. Amer.}, 263(6):84--91, 1990.

\bibitem{Ronneberger2015U}
Olaf Ronneberger, Philipp Fischer, and Thomas Brox.
\newblock U-net: Convolutional networks for biomedical image segmentation.
\newblock In {\em MICCAI}, 2015.

\bibitem{shaham2019singan}
Tamar~Rott Shaham, Tali Dekel, and Tomer Michaeli.
\newblock Singan: Learning a generative model from a single natural image.
\newblock In {\em ICCV}, 2019.

\bibitem{Sharma2018CSGNet}
Gopal Sharma, Rishabh Goyal, Difan Liu, Evangelos Kalogerakis, and Subhransu
  Maji.
\newblock Csgnet: Neural shape parser for constructive solid geometry.
\newblock In {\em CVPR}, 2018.

\bibitem{shocher2019ingan}
Assaf Shocher, Shai Bagon, Phillip Isola, and Michal Irani.
\newblock Ingan: Capturing and remapping the ``dna'' of a natural image.
\newblock In {\em ICCV}, 2019.

\bibitem{shocher2018zero}
Assaf Shocher, Nadav Cohen, and Michal Irani.
\newblock “zero-shot” super-resolution using deep internal learning.
\newblock In {\em CVPR}, 2018.

\bibitem{Teboul2010Segmentation}
Olivier Teboul, Loic Simon, Panagiotis Koutsourakis, and Nikos Paragios.
\newblock Segmentation of building facades using procedural shape priors.
\newblock In {\em CVPR}, 2010.

\bibitem{ulyanov2018deep}
Dmitry Ulyanov, Andrea Vedaldi, and Victor Lempitsky.
\newblock Deep image prior.
\newblock In {\em CVPR}, 2018.

\bibitem{SymmetryHierarchy}
Yanzhen Wang, Kai Xu, Jun Li, Hao Zhang, Ariel Shamir, Ligang Liu, Zhiquan
  Cheng, and Yueshan Xiong.
\newblock Symmetry hierarchy of man-made objects.
\newblock {\em CGF}, 30(2):287--296, 2011.

\bibitem{xie2012image}
Junyuan Xie, Linli Xu, and Enhong Chen.
\newblock Image denoising and inpainting with deep neural networks.
\newblock In {\em NeurIPS}, 2012.

\bibitem{xiong2019foreground}
Wei Xiong, Zhe Lin, Jimei Yang, Xin Lu, Connelly Barnes, and Jiebo Luo.
\newblock Foreground-aware image inpainting.
\newblock In {\em CVPR}, 2019.

\bibitem{yan2018shift}
Zhaoyi Yan, Xiaoming Li, Mu Li, Wangmeng Zuo, and Shiguang Shan.
\newblock Shift-net: Image inpainting via deep feature rearrangement.
\newblock In {\em ECCV}, 2018.

\bibitem{yang2017high}
Chao Yang, Xin Lu, Zhe Lin, Eli Shechtman, Oliver Wang, and Hao Li.
\newblock High-resolution image inpainting using multi-scale neural patch
  synthesis.
\newblock In {\em CVPR}, 2017.

\bibitem{young2019learning}
Halley Young, Osbert Bastani, and Mayur Naik.
\newblock Learning neurosymbolic generative models via program synthesis.
\newblock In {\em ICML}, 2019.

\bibitem{yu2018generative}
Jiahui Yu, Zhe Lin, Jimei Yang, Xiaohui Shen, Xin Lu, and Thomas~S Huang.
\newblock Generative image inpainting with contextual attention.
\newblock In {\em CVPR}, 2018.

\bibitem{Yu2018Free}
Jiahui Yu, Zhe Lin, Jimei Yang, Xiaohui Shen, Xin Lu, and Thomas~S Huang.
\newblock Free-form image inpainting with gated convolution.
\newblock In {\em ICCV}, 2019.

\bibitem{zhou2017places}
Bolei Zhou, Agata Lapedriza, Aditya Khosla, Aude Oliva, and Antonio Torralba.
\newblock Places: A 10 million image database for scene recognition.
\newblock {\em IEEE TPAMI}, 40(6):1452--1464, 2017.

\bibitem{Zhou2018Nonstationary}
Yang Zhou, Zhen Zhu, Xiang Bai, Dani Lischinski, Daniel Cohen-Or, and Hui
  Huang.
\newblock Non-stationary texture synthesis by adversarial expansion.
\newblock {\em SIGGRAPH}, 37(4), 2018.

\bibitem{zontak2011internal}
Maria Zontak and Michal Irani.
\newblock Internal statistics of a single natural image.
\newblock In {\em CVPR}, 2011.

\end{thebibliography}
}

\newpage
\quad
\newpage


\section*{Supplementary material}

# Supplementary Material: Program-Guided Image Manipulators

The supplementary material is organized as follows. In Section A, we first discuss the detailed implementation of PG-IM. We then, in Section B, provide ablation studies evaluating the effects of PG-IM's different modules or designs. Finally, we show more results produced by PG-IM in Section C, and also quantitatively compare them with results by the baselines.

## A. Implementation Details

We discuss the implementation details necessary to reproduce our results. Our code and data are available on the project page: http://pgim.csail.mit.edu/.

### A.1. Repeated Object Detection

We replicated the algorithm developed by Lettry et al. [4]. The algorithm works on nearly-regular images and detects the repeated objects within the image. The model builds on the assumption that features learned from ImageNet [1] encode visual concepts at various levels: from edges to patches to high-level semantic objects. The algorithm takes as input the feature map from a pretrained AlexNet [3] and outputs a set of 2D coordinates representing the centroids of repeated objects. In this section, we explain the algorithm as six-step procedure.

**1)** For each input feature map **f**, it extracts all activated neurons at different locations. Denote the neuron at the 2D position $(i, j)$ being activated by $act(\mathbf{f}_{i,j}) = 1$, which is given by

$$act(\mathbf{f}_{i,j}) = \mathbf{1}\left[\mathbf{f}_{i,j} == \max_{(u,v),|u-i|\leq K,|v-j|\leq K} \mathbf{f}_{u,v}\right],$$

where $\mathbf{1}[\cdot]$ is the indicator function, and $K$ is a hyperparameter for "kernel size." Feature maps at different layers of the AlexNet use different kernel sizes.

**2)** All activated 2D neurons are collected. For each pair of activated neurons from the same feature map, a displacement vector is computed. For simplicity, we use $d^{i,j} = (d_x^{i,j}, d_y^{i,j})$ to index the $j$-th displacement vector from the $i$-th feature map. A majority voting is performed over all displacement vectors. The output of this step is an image-level displacement vector $d^* = (d_x^*, d_y^*)$ between centroids of different objects

$$d_x^* = \arg\max_x \sum_{i,j} \exp(-\|d_x^{i,j} - x\|_2^2) \qquad \text{and} \qquad d_y^* = \arg\max_y \sum_{i,j} \exp(-\|d_y^{i,j} - y\|_2^2).$$

**3)** Based on the image-level displacement vector $d^*$, all displacement vectors are filtered. A vector is selected if

$$\|d^{i,j} - d^*\|_2 \leq 3 \times \alpha,$$

where $\alpha$ is a scalar hyperparameter of radius.

**4)** Each feature map is weighted by the number of selected displacement vectors from this feature map. Only highly weighted feature maps are selected.

**5)** For each 2D coordinate $(x, y)$, we define a helper function

$$\mathcal{M}(x, y) \triangleq (x \bmod d_x^*, y \bmod d_y^*).$$

An offset $o^*$ is then determined by optimizing the following objective:

$$o^* = \arg\min_o \sum_{i,j} \left( w_{i,j} \| M(d^{i,j} - o) - \frac{d^*}{2} \| \right),$$

where

$$w_{i,j} = \frac{1}{K_i + \phi} \cdot \exp\left(-\frac{\|d^{i,j} - d^*\|_2^2}{2\alpha^2}\right).$$

$K_i$ is the number of displacement vectors of feature map $i$. $\phi$ and $\alpha$ are hyperparameters. In our implementation, we use gradient descent to find the optimal $o^*$.

**6)** For each object region detected, the coordinates for all activated neurons within the region vote for the centroid. This step allows distortions of objects that may cause irregular lattices.

### A.2. Adaptive Network Depth for PatchGAN

In order to encourage creative and sharp patch generation, we employ a discriminator loss provided by AdaPatchGAN, our adaptive variant of PatchGAN [2], in addition to the L1 loss. In the original PatchGAN, the classifier architecture is fixed with a receptive field of a certain size. However, in images of our interest, the scale of texture varies violently across different images: the repeated pattern is sometimes a small entity but appearing many times, while other times the pattern is a larger object appearing only a few times. Given a single image, since its inferred program description provides a way of dividing it into patches, AdaPatchGAN computes a rough patch size and automatically adds convolutional layers until its receptive field is large enough to cover this patch size. In practice, we observe that our AdaPatchGAN enables the generator to produce more realistic patches than the original PatchGAN.

### A.3. Extrapolation as Recurrent Inpainting

As mentioned in the main text, a single neural painting network (NPN) trained on the task of image inpainting is able to perform additional manipulation tasks including extrapolation without any finetuning. The straightforward way of doing this is treating the pixels to extrapolate as missing, and just inpainting them in one go. This approach of treating extrapolation as "feedforward inpainting" of multiple objects, however, results in distorted objects that fail to look realistic, as shown in an ablation study below. This is because NPNs are better at inpainting a localized patch (what they are trained to do as in inpainting) than inpainting a region spanning multiple objects.

Bearing this key observation in mind, our NPNs cast extrapolation into a task of *recurrent* inpainting, where it inpaints one object at a time, conditioned on the context provided by the previous inpainting. Note that this is only possible because the NPNs are guided by programs of images and therefore know how many objects should be extrapolated as well as where each of them should appear. Implementation-wise, this is achieved by repeatedly filling the current empty patch with the network output, and running network inference on the next empty patch. By this design, an empty patch gets inpainted so that it connects seamlessly to both the previously inpainted patches and the original image boundary.

### A.4. Detailed Network Specifications and Training Parameters

We append to the end of this document a commented printout of the generator architecture, where channel numbers of the feature maps, kernel sizes, strides, padding, and more can be found. The training of PG-IM generally follows the procedure described in pix2pix [2], except that we do not decay our learning rate after 100 epochs like pix2pix does. Each of our epochs contains 1,000 training samples (i.e., random crops at different scales of the same image of interest), and the networks usually converge within 150 epochs.

## B. Ablation Studies

We report three ablations studies to provide intuitions as to why PG-IM works. Key to our approach is the use of program-like descriptions for images in guiding neural painting networks (NPNs) to perform pixel manipulations. The first study demonstrates why such descriptions facilitate the image manipulation tasks. Next, we justify a crucial design for image extrapolation: extrapolation should be cast into the task of *recurrent* inpainting for NPNs, rather than "feedforward inpainting." Supplemental to Figure 7 of the main paper, the third study emphasizes the advantages of considering high-level attributes in the lower-level image manipulation tasks.

### B.1. With *vs*. Without Source Patches

Our program-like description for an image informs NPNs of where the pattern of interest (repeatedly) appears in the image. NPNs can then align the available patterns to the corrupted or missing one in order to inpaint it (for simplicity, we discuss only image inpainting in this study, but the conclusion stands for the other tasks). These aligned image patches serve as "source patches," from which NPNs smartly copy to fill in the missing pixels.

Without program-like descriptions, NPNs would not have access to these source patches, which in turn leads to inferior performance such as results with the correct background but missing objects. As Figure 1 shows, when programs (and hence source patches) are not available to the networks to smartly copy from, they fail to ensure the presence of the missing object in their inpainting (e.g., the cross in the left pane and the diamond in the right pane). In contrast, our NPNs exploit the source patches provided by programs and can inpaint the missing objects even if they are corrupted completely in the input. This result supports that PG-IM outperforms repetition-aware but "programless" CNNs.

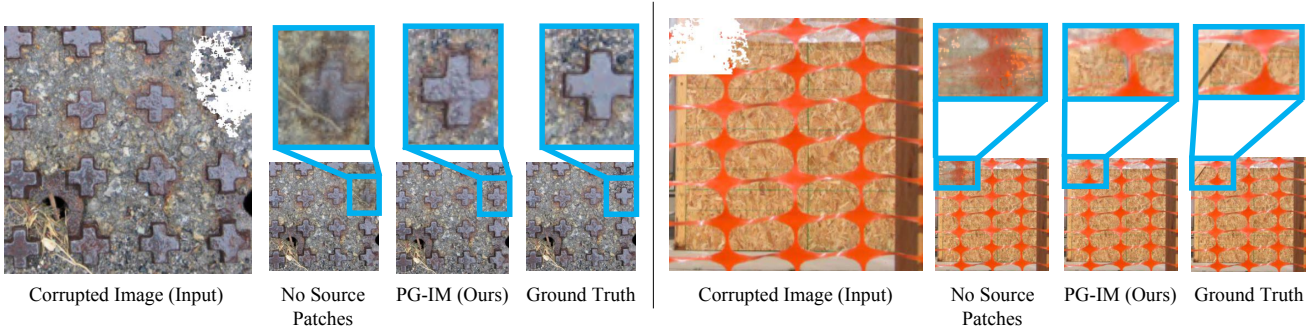

Figure 1: Inpainting with *vs*. without the source patches provided by program. For each pane, the leftmost figure is the corrupted input to inpaint; to the right are results inpainted with and without source patches as well as the ground truth. With programs describing where the object centers are, our NPNs align the intact objects to the corrupted object and exploit such alignment to inpaint the missing pixels. Without such source patches, the networks have difficulty ensuring the "objectness" of the inpainted patch.

## B.2. With *vs*. Without Recurrent Inpainting

In this study, we show empirical evidence that NPNs produce better results when extrapolation is cast into the task of recurrent inpainting than feedforward inpainting. As Figure 2 shows, when the entire stripe gets extrapolated in one go ("extrapolation as feedforward inpainting"), the extrapolated objects tend to be distorted, since our NPNs are trained only on more localized inpainting. By treating extrapolation as *recurrent* inpainting, our NPNs are essentially doing, repeatedly, what they are trained on and hence generate more realistic-looking results.

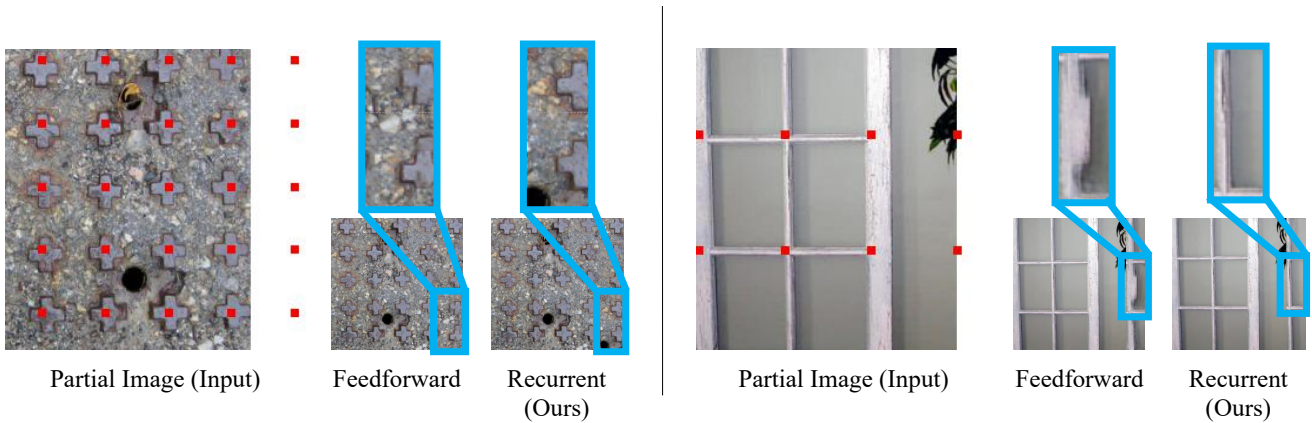

Figure 2: Extrapolation as feedforward *vs*. recurrent inpainting. For each pane, the leftmost figure is the partial input to be extrapolated; to the right are results extrapolated in one go and recurrently. Recall that the networks are trained only on inpainting one entity, so the networks tend to produce distorted objects if they are asked to inpaint the whole stripe spanning multiple entities. Instead, our NPNs treat extrapolation as recurrent inpainting, where they repeatedly solve the task they are good at and therefore produce better results.

## B.3. With *vs*. Without Considering Attributes

In Figure 7 of the main paper, we show our program-like descriptions of images can naturally incorporate attributes, enabling the NPNs to perform attribute-aware image manipulation, e.g., by referencing only patches with the same attributes as the patch to inpaint. We supplement two more examples demonstrating why attribute-aware image manipulations are important, and how PG-IM achieves superior performance on such tasks.

As shown in Figure 3 left, the tomato to be inpainted is next to four orange tomatoes and one green tomato. Since convolutions are more of a local operation, one can expect a convolutional generator to produce a tomato with mixed colors of both orange and green, as is the case for PG-IM w/o Attributes. However, we, as humans, reason at a higher level over the entire image, coming to the conclusions that the tomato to inpaint on the left should be orange, and that the candy to inpaint on the right should be red. With similar capabilities, PG-IM correctly preserves the global regularity of attributes in its inpainting.

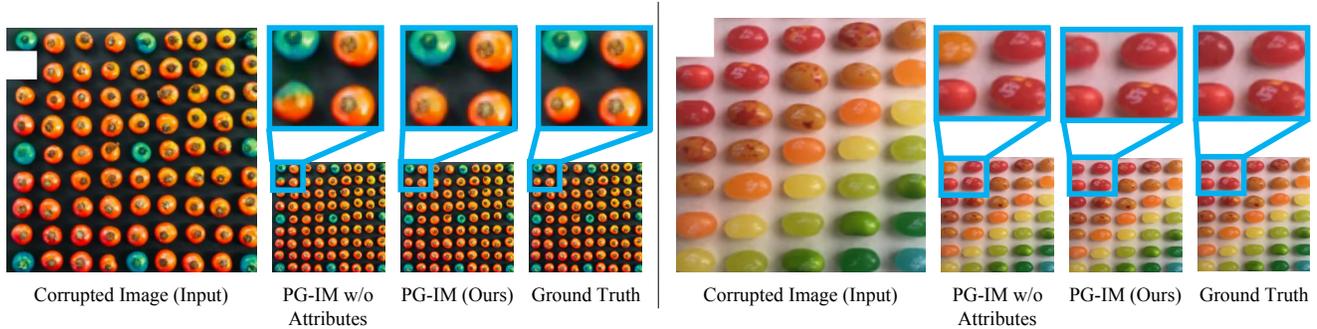

| Corrupted Image (Input) | PG-IM w/o Attributes | PG-IM (Ours) | Ground Truth | Corrupted Image (Input) | PG-IM w/o Attributes | PG-IM (Ours) | Ground Truth |

Figure 3: Inpainting with *vs.* without considering object attributes. For each pane, the leftmost figure is the corrupted input to be inpainted; to the right are results inpainted with and without considering attributes as well as the ground truth. The attribute-agnostic networks tend to produce patches with mixed colors that break the global regularity of attributes. PG-IM is able to respect such regularity by referencing only patches with the same attribute as the patch to inpaint.

## C. More Experiments and Results

We supplement more results by PG-IM on the tasks of image inpainting, extrapolation, and regularity editing. For PatchMatch and Image Quilting, we search for one set of optimal hyperparameters and use that for the entire test set. Similarly, we train Non-Stationary [6] and PG-IM with single sets of optimal hyperparameters on each test image. For GatedConv [5], we use the trained model released by the authors.

### C.1. Inpainting

Figure 4 shows how PG-IM compares in image inpainting with the baselines on another six test images. PG-IM is able to ensure the presence of the repeated object in its inpainting (e.g., the cross structure in Image 1 and the yellow tag in Image 5), while the baselines tend to have the object missing (e.g., PatchMatch completely misses the yellow tag in Image 5) or incomplete (e.g., the pig in Image 3) due to lack of the concept of objects.

### C.2. Extrapolation

We supplement four more examples of image extrapolation by PG-IM and how they compare with results by the baselines. As Figure 5 shows, extrapolation by PG-IM is sharp, respects the global regularity (e.g., the "pig array" in Image 3), and seamlessly connects to the original images. In contrast, PatchMatch tends to blur over the image boundaries for smooth transition to the extrapolated contents (e.g., the blurriness over the boundary in Image 2). Image Quilting and GatedConv tend to break the global regularity (e.g. Images 2 and 4). Finally, Non-Stationary should be considered as a super-resolution method instead of an extrapolation one, because it essentially replicates the patterns on a larger canvas and interpolates in between; notice how it does *not* extrapolate beyond the texture boundaries in Images 2 and 4.

### C.3. Regularity Editing

We supplement four more examples of image regularity editing by PG-IM in Figure 6. Although PG-IM has been trained only on inpainting, with the program descriptions of images, it is naturally capable of exaggerating the irregularity in the images' global structures.

### C.4. Failure Cases

Figure 7 demonstrates a failure case for inpainting. In this challenging case, the transmission and reflection effects make the repetitive patterns less homogeneous, leading to non-photorealistic inpainting.

### C.5. Baseline Finetuning

PG-IM learns from a single input image by exploiting the repetition structure, whereas other learning-based methods (GatedConv and PartialConv) learn from a large dataset (Places365). We supplement extra experiments, where we finetune the GatedConv model (pre-trained on Places365) on 50 Facade images and evaluate it on the held-out images (Figure 8). Learning from similar facade images does help the model produce more photorealistic results (Inception score: $1.187 \to 1.191$). However, the model still fails on structured objects such as windows, which can be well handled by our patch-based, program-guided NPN (PG-IM's Inception score is 1.210).

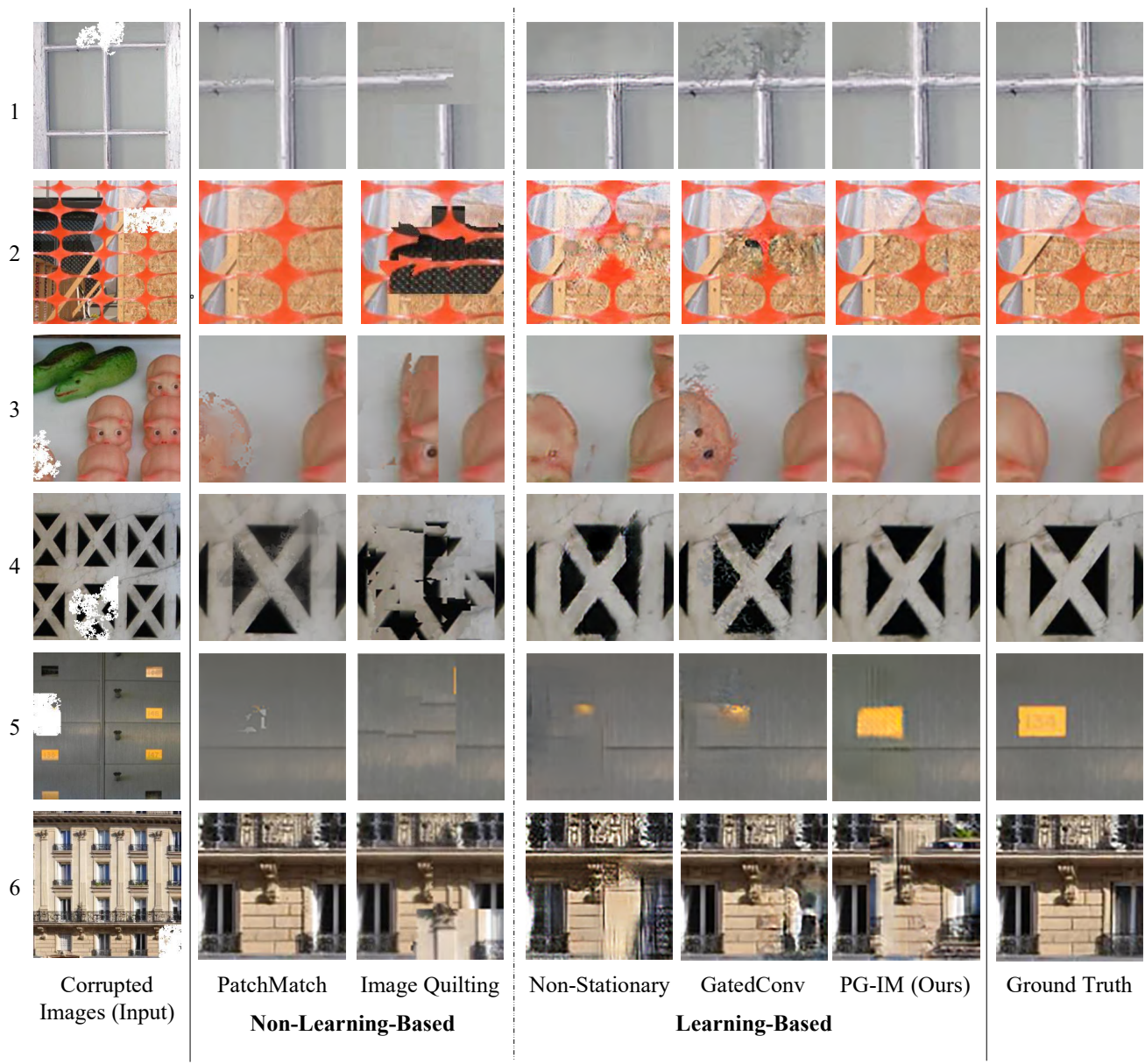

Figure 4: More inpainting results by PG-IM and the baselines. The white pixels in the leftmost column are missing pixels to inpaint; the rightmost column shows the ground-truth patches. For easier comparison, we show only the close-up views of the inpainted region (and its proximity for some context). Aware of the images' global regularity, PG-IM inpaints the missing objects that maintain this regularity. Note this is not necessarily the case for other methods: baselines miss out the pig entity of Image 3 and the yellow tag of Image 5. Although PatchMatch gets the cross structure in Image 1, its result is less realistic than ours. Another advantage of PG-IM over the baselines is the ability to generate pixels that seamlessly connect to the original image contents (compare the results for Images 1, 2, and 4).

## References

[1] Jia Deng, Wei Dong, Richard Socher, Li-Jia Li, Kai Li, and Li Fei-Fei. Imagenet: A large-scale hierarchical image database. In *CVPR*, 2009. 1

[2] Phillip Isola, Jun-Yan Zhu, Tinghui Zhou, and Alexei A Efros. Image-to-image translation with conditional adversarial networks. In *CVPR*, 2017. 2

[3] Alex Krizhevsky, Ilya Sutskever, and Geoffrey E Hinton. Imagenet classification with deep convolutional neural networks. In *NeurIPS*, 2012. 1

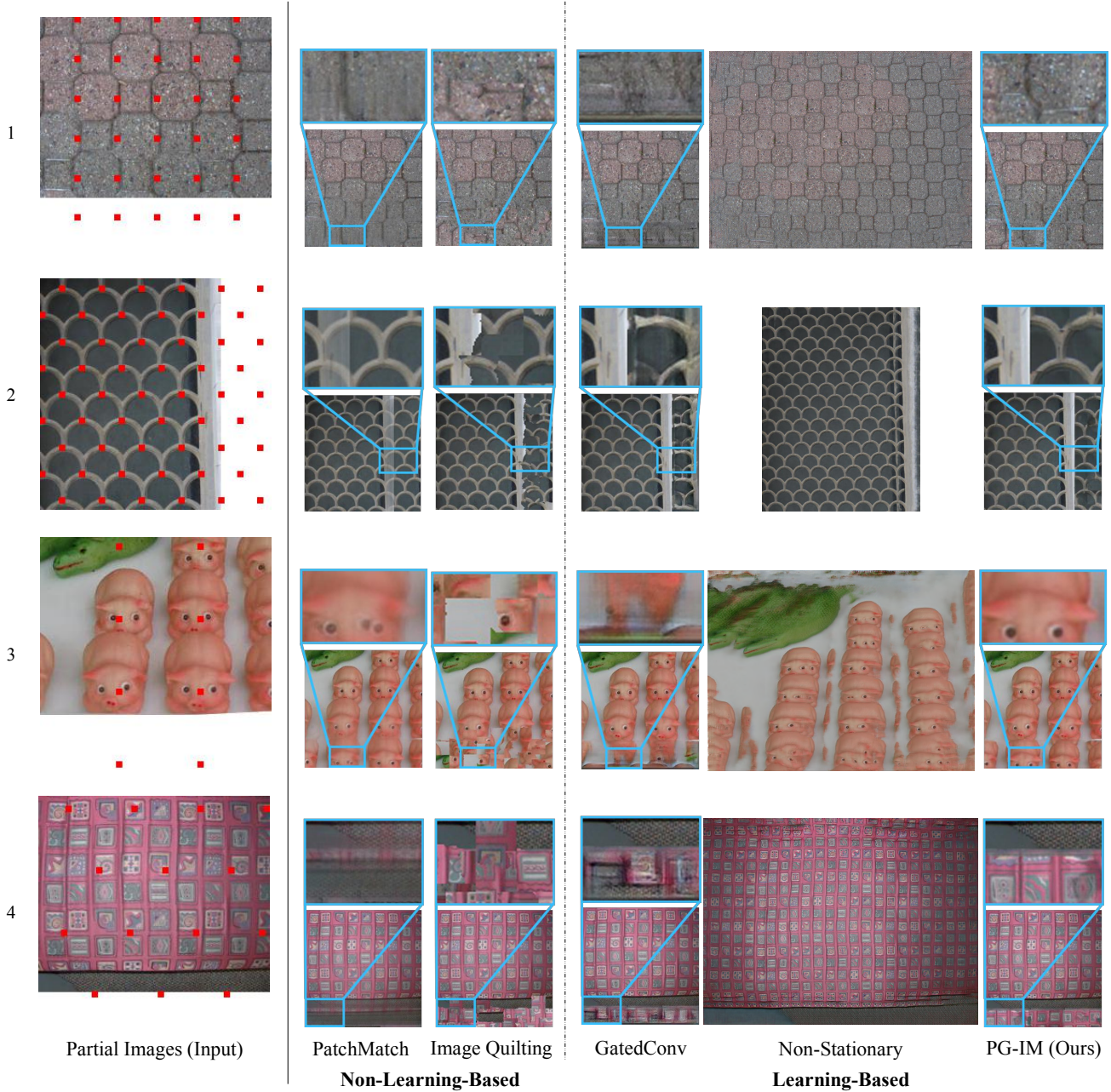

Figure 5: Extrapolation results by PG-IM and the baselines. The white pixels in the leftmost column indicate the pixels to be extrapolated, and the red dots are object centers given by the program descriptions. With such program descriptions, PG-IM knows where to extrapolate to automatically, whereas other methods require the user to specify where to extrapolate to. PG-IM generates realistic images while preserving the global regularity. In contrast, GatedConv fails to capture the regularity; Non-Stationary does not preserve the original content of images; non-learning-based baselines sometimes generate blurry images because of the repetition of similar objects in the partial image.

[4] Louis Lettry, Michal Perdoch, Kenneth Vanhoey, and Luc Van Gool. Repeated pattern detection using cnn activations. In *WACV*, 2017. 1

[5] Jiahui Yu, Zhe Lin, Jimei Yang, Xiaohui Shen, Xin Lu, and Thomas S Huang. Free-form image inpainting with gated convolution. In *ICCV*, 2019. 4

[6] Yang Zhou, Zhen Zhu, Xiang Bai, Dani Lischinski, Daniel Cohen-Or, and Hui Huang. Non-stationary texture synthesis by adversarial expansion. *SIGGRAPH*, 37(4), 2018. 4

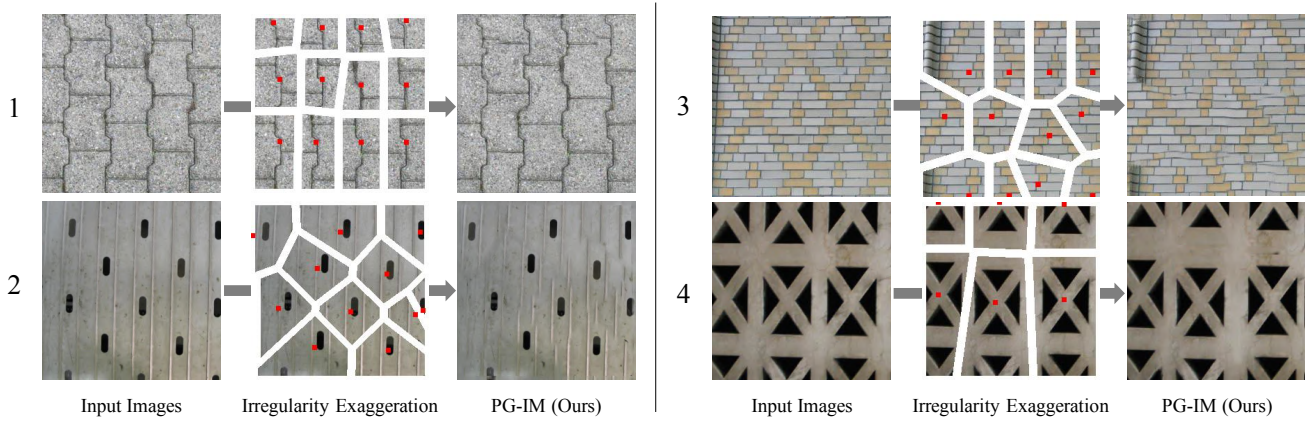

Figure 6: PG-IM enables automated and semantic-aware irregularity exaggeration. By comparing the centroids of the detected objects and the ones reconstructed by the program, we can measure and exaggerate the structural irregularity of input images. In these examples, we first multiply by 2 the displacement vectors between the object centroids provided by the programs and the detected object centroids, and then randomly flip the sign for each displacement vector. According to these new displacement vectors, we shift the patches and then let the NPN fill in the missing pixels.

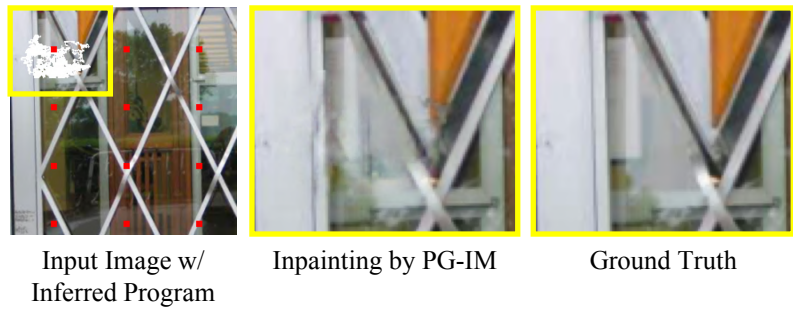

Figure 7: A failure case.

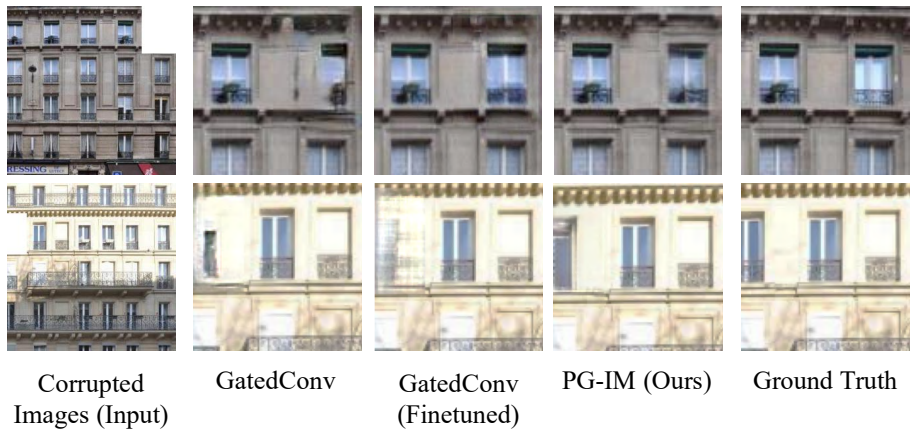

Figure 8: GatedConv model finetuned on the Facade dataset. Note that our PG-IM is only trained on a single input image.

# Appendix: Detailed Generator Architecture

```
('1', ReppadConv2d(3, 64, 3, stride=1, padding=1)),
('1a', nn.ELU()),
# 2: maxpool across tracks
# 3: concat 2
('4:+pool', ReppadConv2d(128, 96, 1, stride=1)),
('4a', nn.ELU()),
('5', ReppadConv2d(96, 96, 4, stride=2, padding=1)),
('5a', nn.ELU()),
# 6: maxpool across tracks
# 7: concat 6
('8:+pool', ReppadConv2d(192, 128, 1, stride=1)),
('8a', nn.ELU()),
('9', ReppadConv2d(128, 128, 4, stride=2, padding=1)),
('9a', nn.ELU()),
# 10: maxpool across tracks
# 11: concat 10
('12:+pool', ReppadConv2d(256, 256, 1, stride=1)),
('12a', nn.ELU()),
('13', ReppadConv2d(256, 256, 4, stride=2, padding=1)),
('13a', nn.ELU()),
# 14: maxpool across tracks
# 15: concat 14
('16:+pool', ReppadConv2d(512, 384, 1, stride=1)),
('16a', nn.ELU()),
('17', ReppadConv2d(384, 384, 4, stride=2, padding=1)),
('17a', nn.ELU()),
('18', ReppadConv2d(384, 384, 3, stride=1, padding=1)),
('18a', nn.ELU()),
# ----------------
('19', nn.ConvTranspose2d(384, 384, 4, stride=2, padding=1)),
('19a', nn.ELU()),
# 20: maxpool across tracks
# 21: concat 16a, 20
('22:+pool+16a', ReppadConv2d(1152, 384, 1, stride=1)),
('22a', nn.ELU()),
('23', ReppadConv2d(384, 384, 3, stride=1, padding=1)),
('23a', nn.ELU()),
('24', nn.ConvTranspose2d(384, 256, 4, stride=2, padding=1)),
('24a', nn.ELU()),
# 25: maxpool across tracks
# 26: concat 12a, 25
('27:+pool+12a', ReppadConv2d(768, 256, 1, stride=1)),
('27a', nn.ELU()),
('28', ReppadConv2d(256, 256, 3, stride=1, padding=1)),
('28a', nn.ELU()),
('29', nn.ConvTranspose2d(256, 192, 4, stride=2, padding=1)),
('29a', nn.ELU()),
# 30: maxpool across tracks
# 31: concat 8a, 30
('32:+pool+8a', ReppadConv2d(512, 192, 1, stride=1)),
('32a', nn.ELU()),
('33', ReppadConv2d(192, 192, 3, stride=1, padding=1)),
('33a', nn.ELU()),
('34', nn.ConvTranspose2d(192, 96, 4, stride=2, padding=1)),
('34a', nn.ELU()),
# 35: maxpool across tracks
```

```python
# 36: concat 1a, 35
('37:+pool+1a', ReppadConv2d(256, 96, 1, stride=1)),
('37a', nn.ELU()),
('38', ReppadConv2d(96, 96, 3, stride=1, padding=1)),
('38a', nn.ELU()),
# 39: maxpool across tracks
('40:pool', ReppadConv2d(96, 64, 3, stride=1, padding=1)),
('40a', nn.ELU()),
('41', ReppadConv2d(64, nch_out, 3, stride=1, padding=1)),
```



\end{document}